\documentclass[accepted]{uai2024}

\usepackage[american]{babel}
\usepackage{hyperref}
\usepackage{graphicx} %
\usepackage{amsthm} %
\usepackage{amsmath} %
\usepackage{amsfonts} %
\usepackage{amssymb} %
\usepackage{bm} %
\usepackage{soul} %
\usepackage{scalerel} %
\usepackage{mathtools} %
\usepackage{booktabs} %
\usepackage{tikz} %
\usepackage{caption} %
\usepackage{xcolor} %
\usepackage{centernot} %
\usepackage{subcaption} %
\usepackage[ruled]{algorithm2e} %
\usepackage{multirow} %
\usepackage{float} %

\usetikzlibrary{positioning}
\usetikzlibrary{arrows.meta}
\usetikzlibrary{calc}
\usetikzlibrary{shapes.geometric}
\usetikzlibrary{fit}
\usetikzlibrary{decorations.pathreplacing}

\usepackage{natbib} %
\usepackage{amsmath}
\usepackage{amsthm}
\usepackage{bm}
\usepackage{cleveref}
\usepackage{tcolorbox}

\newcounter{issuecount}
\newcounter{todocount}
\newcounter{notecount}
\renewcommand{\vec}[1]{\ensuremath\bm{{#1}}}

\DeclareMathOperator*{\argmin}{arg\,min}

\newtheorem{definition}{Definition}
\newtheorem{lemma}{Lemma}

\newtheorem{theorem}{Theorem}
\newtheorem{corollary}{Corollary}
\newtheorem{example}{Example}

\newcommand{\anc}{\dashrightarrow}

\DeclarePairedDelimiter\parentheses{\lparen}{\rparen}
\newcommand{\pa}[1]{\operatorname{Pa} \parentheses*{#1}}
\newcommand{\an}[1]{\operatorname{An} \parentheses*{#1}}

\newcommand{\scm}[1]{\ensuremath\mathcal{#1}}

\newcommand{\rst}[1]{\operatorname{Rst}\left(#1\right)}

\newcommand{\taudirect}{{\xrightarrow{\mbox{\tiny $\mat{T}$}}}}
\newcommand{\rel}[1]{\ensuremath{\Pi_R(#1)}}

\newcommand{\real}{\ensuremath\mathbb{R}}

\newcommand{\set}[1]{\ensuremath\bm{{#1}}}
\newcommand{\setdef}[1]{\ensuremath\left\{{#1}\right\}}
\newcommand{\dom}[1]{\ensuremath{\mathcal{D}(#1)}}

\newcommand{\dset}{\ensuremath{\mathcal{D}}}

\newcommand{\dist}[1]{\ensuremath\mathbb{{#1}}}
\newcommand{\indep}{\perp\!\!\!\perp}

\newcommand{\mat}[1]{{\ensuremath\boldsymbol{\mathbf{#1}}}}

\newcommand{\tr}{\ensuremath{\top}}
\newcommand{\inv}{\ensuremath{{-1}}}

\title{%
Learning Causal Abstractions of Linear Structural Causal Models
}

\author[1]{\href{mailto:<riccardo.massidda@phd.unipi.it>?Subject=Causal Models and Stuff}{Riccardo Massidda}}
\author[2]{Sara Magliacane}
\author[1]{Davide Bacciu}
\affil[1]{%
  Department of Computer Science\\
  Università di Pisa\\
  Pisa, IT
}
\affil[2]{%
  Informatics Institute\\
  University of Amsterdam\\
  Amsterdam, NL
}

\begin{document}

\maketitle

\begin{abstract}
The need for modelling causal knowledge at different levels of granularity arises in several settings. Causal Abstraction provides a framework for formalizing this problem by relating two Structural Causal Models at different levels of detail. Despite increasing interest in applying causal abstraction, e.g. in the interpretability of large machine learning models, the graphical and parametrical conditions under which a causal model can abstract another are not known. Furthermore, learning causal abstractions from data is still an open problem. In this work, we tackle both issues for linear causal models with linear abstraction functions. First, we characterize how the low-level coefficients and the abstraction function determine the high-level coefficients and how the high-level model constrains the causal ordering of low-level variables. Then, we apply our theoretical results to learn high-level and low-level causal models and their abstraction function from observational data. In particular, we introduce Abs-LiNGAM, a method that leverages the constraints induced by the learned high-level model and the abstraction function to speedup the recovery of the larger low-level model, under the assumption of non-Gaussian noise terms. In simulated settings, we show the effectiveness of learning causal abstractions from data and the potential of our method in improving scalability of causal discovery.
\end{abstract}

\begin{figure*}[t]
  \centering
  \begin{subfigure}[b]{0.48\textwidth}
    \includegraphics[width=\linewidth]{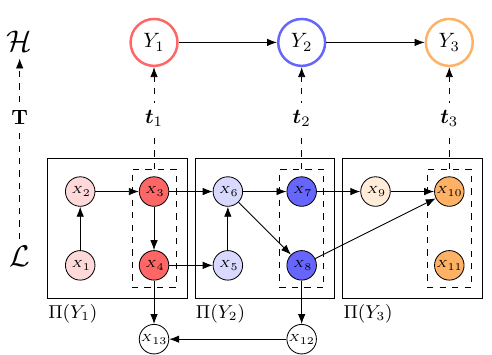}
    \caption{%
      $\mat{T}$-Abstraction
    }\label{subfig:a}
  \end{subfigure}
  \hfill
  \begin{subfigure}[b]{0.48\textwidth}
    \includegraphics[width=\linewidth]{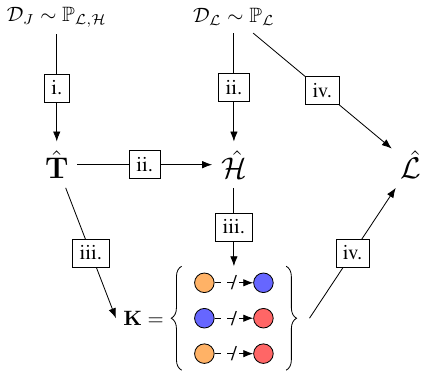}
    \caption{
      Abs-LiNGAM
    }\label{subfig:b}
  \end{subfigure}
  \caption{%
    An overview of our contributions:
    (a.) A linear SCM~$\mathcal{H}$, representing the \emph{abstract} causal model,
    is a $\mat{T}$-abstraction
    of a linear SCM~$\mathcal{L}$, representing the \emph{concrete} causal model,
    whenever the
    linear transformation~$\mat{T}$
    from
    concrete
    to abstract variables
    is interventionally consistent, i.e., whenever it relates
    both values and
    interventions on the abstract model and the concrete model.
    We prove
    that,
    for each abstract variable~$Y$,
    the transformation $\mat{T}$
    induces
    a block~$\Pi(Y)$ of concrete causal variables
    that necessarily follows
    the causal ordering
    of the abstract model
    and whose parameters
    are constrained
    by the abstract coefficients.
    For each block,
    the abstraction function
    depends on a possibly smaller
    subset of \emph{relevant} variables,
    which we portray as dashed.
    (b.) We
    propose
    Abs-LiNGAM,
    a method
    to speedup the causal discovery
    of the concrete model~$\mathcal{L}$
    given an additional dataset $\dset_{J}$
    sampled from the joint distribution
    of the abstract and the concrete model.
    In order,
    Abs-LiNGAM
    (i.) reconstructs the transformation $\mat{T}$,
    (ii.) fits the abstract model by abstracting the concrete dataset $\dset_{\scm{L}}$,
    (iii.) infers a set of constraints $\mat{K}$ on which paths cannot exist in the concrete graph,
    and finally (iv.) discovers the concrete model in a search space reduced by the constraints.
  }\label{fig:mainfig}
\end{figure*}

\section{Introduction}

Causal Abstraction formalizes the property of distinct causal models to describe the same phenomenon with different levels of detail~\citep{beckers2019abstracting}. Despite having different variables and mechanisms, whenever two Structural Causal Models (SCMs) are in an abstraction relation, there must always exist at least one implementation on the low-level \emph{concrete} model of any property of the high-level \emph{abstract} one --- such as values, interventions, mechanisms, and endogenous or exogenous distributions. 

Abstract causal models allow the interpretation of causal models with large number of variables, such as in climate phenomena~\citep{chalupka2016unsupervised} or brain activation patterns~\citep{dubois2020personality}. Causal Abstraction has also found wide interest in explainable AI to align machine representations with human-interpretable concepts in feedforward neural networks~\citep{geiger2021causal,geiger2023causal}, concept-based neural networks~\citep{marconato2023interpretability}, and Large Language Models~\citep{wu2024interpretability,geiger2024finding}.

Previous works on the definition of Causal Abstraction do not focus on the graphical and parametrical conditions for two models to be in an abstraction relation. Furthermore, the problem of learning abstractions from data, when the high-level model is not known, is still open. In this context, \citet{zennaro2022abstraction} and \citet{geiger2023causal} propose methods to learn an abstraction function assuming to know both the low-level and the abstract model, while \citet{chalupka2016unsupervised,kekic2023targeted} and \citet{felekis2024causal} assume to have at least the graphical structure of the abstract model.

In this paper, we tackle these issues by focusing on the scenario where two linear SCMs are abstracted by a linear transformation, as shown in \Cref{fig:mainfig}.
In particular,
we study necessary and sufficient conditions
for abstraction
in terms of the edges
and the coefficients of the models.
We then propose Abs-LiNGAM,
a strategy
to learn from data
the abstract model,
the concrete model,
and their abstraction function
under the further assumption
of non-Gaussian exogenous noise.
We summarize our contributions as follows:

\begin{enumerate}
  \item%
    We first prove that abstract edges
    necessarily require
    edges in the low-level model
    to connect relevant variables,
    i.e., variables on which
    the abstraction function
    directly depends (\Cref{theo:connectivity}).
    Then,
    we show that the abstraction necessarily arranges concrete variables in adjacent and disjoint blocks that must follow the abstract causal ordering (\Cref{theorem:absord}).
  \item%
    We then prove a necessary and sufficient condition
    for causal abstraction
    that relates
    the coefficients
    of the linear models and
    the abstraction function (\Cref{theo:concretization}).
    In this way,
    we can characterize
    the set of all concrete models
    that are abstracted by a given abstract SCM
    and design a complete and correct algorithm to sample any model from this set (\Cref{alg:samplingblocks}).
  \item%
    We introduce Abs-LiNGAM,
    a method to speedup the causal discovery
    of large linear non-Gaussian models
    given an additional and small dataset sampled
    from the observational joint distribution
    of the model
    and one of its abstractions.
    Abs-LiNGAM recovers the abstraction function,
    learns the abstract model
    using low-level data,
    and finally
    constrains the recovery
    of the concrete model
    by ensuring that the necessary conditions
    we introduced
    are satisfied (\Cref{alg:abslingam}).
    \item%
    As we report in \Cref{sec:experiments},
    experiments in simulated settings
    show that Abs-LiNGAM substantially reduces the search space, and thus the execution time, compared to directly solving
    the problem on the low-level dataset
    with DirectLiNGAM~\citep{shimizu2011directlingam}.
\end{enumerate}

We also publicly release online
the code 
of Abs-LiNGAM and the experimental settings\footnote{\url{https://github.com/rmassidda/causabs}}.

\section{Background}\label{sec:background}

Given a set of variables~$\set{X}$,
we denote the domain
of each variable~${X\in\set{X}}$
as~$\dom{X}$ 
and of any subset ${\set{V}\subseteq\set{X}}$
as~${\dom{\set{V}}}$.
We
define
a Structural Causal Model~\citep[SCM;][]{pearl2009causality}
as a tuple ${\scm{M} =
\left(
  \set{X},
  \set{E},
{\{f_X\}}_{X\in\set{X}},
\dist{P}_{\set{E}}\right)}$,
where
\begin{enumerate}[nosep]
  \item%
    $\set{X}$ is a set containing $d$ distinct \emph{endogenous} variables,
  \item%
    $\set{E}$ is a set containing $d$ distinct \emph{exogenous} variables,
  \item%
    ${f_X\colon\dom{\pa{X}\cup\{E_X\}}\to\dom{X}}$
    is a \emph{causal mechanism}, i.e. a function
    that determines the value
    of the variable $X\in\set{X}$
    given its parents $\pa{X}$
    and the exogenous noise term $E_X\in\set{E}$,
  \item%
    $\dist{P}_{\set{E}}$ is the joint distribution over $\set{E}$.
\end{enumerate}
We assume
that parental relations
define
a directed acyclic graph $\scm{G_M}$
and, consequently,
that the reduced form of the model
always has a unique solution~\citep{bongers2021foundations}.
By slightly abusing the notation,
we denote as~$\scm{M}$
both the SCM
and its reduced form $\scm{M}\colon\dom{\set{E}}\to\dom{\set{X}}$
mapping exogenous to endogenous values.
A hard intervention
is an assignment $i = (\set{V}\gets\vec{v})$
on a subset of variables $\set{V}\subseteq\set{X}$
that replaces
each mechanism of the variables $\set{V}$
with a constant value $\vec{v}\in\dom{\set{V}}$.
We denote as $\set{I}^\ast$
the set of all hard interventions
on an SCM, containg all possible assignments
to any subset of endogenous variables,
also including the empty intervention.
Formally,
an intervention~$i$
results in a different SCM $\scm{M}^i$
defined by
the tuple
${\left(\set{X},\set{E},
{\{f_X^i\}}_{X\in\set{X}},
\dist{P}_{\set{E}}\right)}$,
where
${f_X^i = f_X}$
if $X\notin\set{V}$
and
$f_X^i(\cdot) = v_X$
otherwise.
We then define
the restriction of an intervened causal model
as the set of values that the model can take
after the intervention,
which we denote as
\begin{align}
  \rst{\scm{M}^{\set{V}\gets\vec{v}}}
  = \setdef{\vec{x}\in\dom{\set{X}}\mid\vec{x}_{\set{V}} = \vec{v}},
\end{align}
where $\rst{\scm{M}}=\dom{\set{X}}$
for a non-intervened SCM.

We assume
faithfulness
and causal sufficiency,
i.e., the absence of hidden confounding and selection bias~\citep{spirtes2000causation}.
In particular,
faithfulness
implies the absence of
canceling paths across variables,
while causal sufficiency
implies mutual independence
of exogenous terms, as in
$E_1 \indep E_2$ for any $E_1,E_2\in\set{E}$.

A linear SCM~${\scm{M}=\left(\set{X}, \set{E}, \mat{W}, \dist{P}_{\set{E}}\right)}$,
also known as a linear Additive Noise Model (ANM)~\citep{peters2017elements},
is an SCM
whose structural equations
are linear
and represented
by an upper-triangular adjacency matrix ${\mat{W}\in\real^{d \times d}}$,
as in
\begin{align}
  \mat{X} = \mat{W}^\tr \mat{X} + \mat{E}.
\end{align}
We can compute the reduced form of the model
in closed form
as,
\begin{align}
  \scm{M}(\vec{e})
  =
  \mat{F}^\tr \vec{e},
\end{align}
where $\mat{F} = {(\mat{I}-\mat{W})}^\inv$ is a $d\times d$ linear transformation.

Causal Abstraction theory
relates variables
across
different SCMs
to determine
whether they
represent
in a \emph{consistent} way
the same system
at different levels of detail~\citep{beckers2019abstracting}.
Overall, we refer to
\emph{concrete}, or low-level, causal models
as $\scm{L}=(\set{X},\set{E},\set{f},\dist{P}_{\set{E}})$
and to
\emph{abstract}, or high-level, causal models
as $\scm{H}=(\set{Y},\set{U},\set{g},\dist{P}_{\set{U}})$,
where ${|\set{X}| \geq |\set{Y}|}$.
An abstraction
requires to consider
two subsets of
allowed
interventions
$\set{I}\subseteq\set{I}^\ast$
and
$\set{J}\subseteq\set{J}^\ast$
respectively
on the concrete
and the abstract model.

In this work,
we focus on \emph{strong} abstractions,
where
any concrete
or abstract intervention
is allowed, i.e.,
$\set{I}=\set{I}^\ast$
and
$\set{J}=\set{J}^\ast$.
Then,
given a surjective function
$\tau\colon\dom{\set{X}}\to\dom{\set{Y}}$,
$\scm{H}$
is a {$\tau$-abstraction}
of $\scm{L}$
if and only if
there exists a surjective function $\gamma\colon\dom{\set{E}}\to\dom{\set{U}}$
on the exogenous variables
such that,
for any low-level intervention $i\in\set{I}$
and
any exogenous configuration
$\vec{e}\in\dom{\set{E}}$,
it holds
\begin{align}\label{eq:intconst}
  \tau(\scm{L}^i(\vec{e}))
  =
  \scm{H}^{\omega(i)}(\gamma(\vec{e})),
\end{align}
where
the intervention map~$\omega\colon\set{I}\to\set{J}$
is uniquely induced by the value abstraction function~$\tau$~\citep{massidda2023causal}.
Formally, $\omega(i)=j$
if and only if
$\rst{\scm{H}^j} = \tau(\rst{\scm{L}^i})$
and $j\in\set{J}$,
otherwise $\omega(i)$ is undefined.
We refer to \Cref{eq:intconst} as the \emph{interventional consistency} property and to the function $\gamma$ as the \emph{exogenous abstraction function}.
Similarly,
we refer to the equation 
\begin{align}
    \tau(\scm{L}(\vec{e}))
    =
    \scm{H}(\gamma(\vec{e}))
\end{align}
on non-intervened models
as \emph{observational consistency}.
Since
the empty intervention
is necessarily a fixed point
of the intervention map~\citep{massidda2023causal},
interventional consistency
implies
observational consistency.

Finally,
as for the causal mechanisms,
we assume that the abstraction function
does not yield cancelling paths
towards abstract variables.
Formally,
the composition
of the abstraction function~$\tau$,
with the concrete model~$\scm{L}$
must not cancel the effect
of concrete variables~$\set{X}$
on abstract variables~$\set{Y}$.

\section{Theory of Linear Causal Abstraction}\label{sec:linabs}

In this section,
we study
graphical
and structural
properties
of linear causal models
in a linear abstraction relation.
First,
we prove that
the set of concrete \emph{relevant} variables
on which each abstract variable
depends
are necessarily disjoint
(\Cref{subsec:linabs}).
Further,
we prove
necessary and sufficient conditions
on the existence of an abstract edge
in terms of the
directed paths
between relevant variables
in the concrete graph
(\Cref{subsec:topabs}).
We then show
that the abstraction function
constrains the causal ordering
of \emph{concrete blocks}
composed of both
relevant and
non-relevant variables
(\Cref{subsec:varcnc}).
Finally,
given the notion of concrete block,
we provide an equivalent
formulation of abstraction
based on the model parameters,
which also 
characterizes
the set of possible
concretizations of an abstract model
(\Cref{subsec:weightconc}).

\subsection{Linear Causal Abstraction}\label{subsec:linabs}

We focus on the scenario
where the value abstraction function~$\tau$
between
an abstract
and a concrete
causal model
is a linear transformation
represented by a matrix~$\mat{T}$.
We
then define
such
linear relation between SCMs
as $\mat{T}$-abstraction.
\begin{definition}[$\mat{T}$-Abstraction]
  Let $\scm{H}$ be a strong $\tau$-abstraction of $\scm{L}$,
  where $\scm{H}$ and $\scm{L}$
  are two SCMs
  respectively on variables $\set{Y}$ and $\set{X}$.
  Then,
  $\scm{H}$ is a $\mat{T}$-abstraction of $\scm{L}$
  whenever
  there exists a linear transformation $\mat{T}\in\real^{d \times b}$,
  where $d=|\set{X}|$ and $b=|\set{Y}|$,
  such that $\tau(\vec{x}) = \mat{T}^\tr \vec{x}$.
\end{definition}

One of the common aspects
of causal abstraction
consists of reducing the dimensionality
of a causal model
by selecting \emph{relevant}
and discarding \emph{irrelevant} variables~\citep{zennaro2022abstraction}.
Therefore,
for each abstract variable~$Y$,
we define its set
of \emph{relevant} variables~$\Pi_R(Y)$
as the set of concrete variables on which it
directly depends
according to the $\mat{T}$-abstraction.
Overall, we refer
to the set of relevant variables
in the concrete model
as the union of the
relevant variables
of each abstract variable
and
to all the remaining as
\emph{irrelevant}.
\begin{definition}[Relevant Variables]\label{def:relvar}
  Let $\scm{H}$ be a $\mat{T}$-abstraction of $\scm{L}$,
  where $\scm{H}$ and $\scm{L}$
  are two SCMs
  respectively on variables $\set{Y}$ and $\set{X}$.
  We define
  the set of relevant concrete variables
  of an abstract variable $Y_j\in\set{Y}$
  as the subset
  \begin{align}
    \Pi_R(Y_j) = \{ X_i \in\set{X} \mid t_{ij} \neq 0 \},
  \end{align}
  where $t_{ij}$
  is the $i$-th element
  on the $j$-th column
  of $\mat{T}$.
  Moreover, we define the set of relevant variables $\Pi_R(\set{Y})$ of the abstract model~$\scm{H}$ as the union of all relevant sets for each variable $Y \in \set{Y}$. Formally,
  \begin{align}
   \Pi_R (\set{Y}) = \bigcup_{Y \in \set{Y}} \Pi_R(Y).
  \end{align} 
  We define as irrelevant the remaining variables in the concrete model $\scm{L}$, i.e. $\set{X} \setminus \Pi_R (\set{Y})$.
\end{definition}

To guarantee surjectivity,
the transformation~$\mat{T}$
must have full column-rank
and,
consequently,
the set of relevant variables
for each abstract variable
must be non empty.
Since we require consistency
to hold
on the set of all possible
abstract hard interventions,
we can easily prove that this implies
that the sets of relevant variables
must be mutually disjoint.
\begin{lemma}[Disjoint Relevant]\label{lemma:disjointrelevant}
  Let $\scm{H}$ be a $\mat{T}$-abstraction of $\scm{L}$,
  where $\scm{H}$ and $\scm{L}$
  are two linear SCMs
  respectively on variables $\set{Y}$ and $\set{X}$.
  Then,
  for any pair of distinct abstract variables $Y_1, Y_2\in\set{Y}$,
  it holds that $\Pi_R(Y_1)\cap\Pi_R(Y_2) = \emptyset$,
  where $\Pi_R(Y_1) \neq \emptyset$ and $\Pi_R(Y_2) \neq \emptyset$.
\end{lemma}
\begin{proof}
    We report the proof in \Cref{proof:disjointrelevant}.
\end{proof}
\citet{beckers2019abstracting}
define \emph{constructive} abstraction
as a special case
where abstract variables
depend on disjoint sets
of low-level variables
and
conjectures
that, under further assumptions,
strong abstraction
might entail constructive abstraction.
Notably,
with our linearity assumptions,
such conjecture
immediately derives from \Cref{lemma:disjointrelevant}.
\begin{corollary}[Constructive Abstraction]\label{cor:constructive}
    Let $\scm{H}$
    be a strong $\tau$-abstraction
    of $\scm{L}$ where
    $\scm{H}$ and $\scm{L}$ are linear SCMs
    and
    $\tau$ is a linear transformation.
    Then,
    $\scm{H}$ is a constructive $\tau$-abstraction
    of $\scm{L}$.
\end{corollary}
\begin{proof}
    We report the proof in \Cref{proof:constructive}.
\end{proof}

\subsection{Graphical Characterization of T-Abstract Linear Causal Models}\label{subsec:topabs}

To characterize the relation
between abstract edges
and the concrete graph
in a $\mat{T}$-abstraction,
we must take into account
that directed paths
between \emph{relevant} variables
in the concrete graph
might be mediated by \emph{irrelevant} variables.
Therefore,
to study
how causal effect propagates,
we say that
a directed path between two relevant variables
is $\mat{T}$-direct
whenever it is mediated
by irrelevant variables only. We denote edges and directed paths between two variables $X_1, X_2$ respectively as $X_1 \to X_2$ and $X_1 \anc X_2$.

\begin{definition}[$\mat{T}$-direct Path]
  Let $\scm{H}$ be a $\mat{T}$-abstraction of $\scm{L}$,
  where $\scm{H}$ and $\scm{L}$
  are two SCMs
  respectively on variables $\set{Y}$ and $\set{X}$ with graphs $\mathcal{G}_\scm{H}$ and $\mathcal{G}_\scm{L}$.
  Given two concrete variables $X_1,X_2\in\set{X}$,
  we say that there exists $\mat{T}$-direct path in $\mathcal{G}_\scm{L}$, denoted as
  $X_1 \taudirect X_2$,
  if and only if
  there exists a directed path $X_1 \anc X_2$ in $\mathcal{G}_\scm{L}$
  such that any other variable $X_3$ in the path
  is irrelevant, i.e.,
  for any abstract variable $Y$,
  it holds $X_3 \not\in\Pi_R(Y)$.
\end{definition}

First,
we show that a $\mat{T}$-direct path
between relevant variables
in the concrete graph
is a sufficient condition
for the presence of an edge
between their corresponding
abstract variables.
Further, as an immediate corollary,
a direct path between relevant variables
entails an abstract direct path.
\begin{lemma}[Sufficient Abstract Connectivity]\label{lemma:sufcon}
  Let $\scm{H}$ be a $\mat{T}$-abstraction of $\scm{L}$,
  where $\scm{H}$ and $\scm{L}$
  are two linear SCMs
  respectively on variables $\set{Y}$ and $\set{X}$ with graphs $\mathcal{G}_\scm{H}$ and $\mathcal{G}_\scm{L}$.
  Then, 
  for any pair
  of relevant variables $X_1,X_2\in\Pi_R(\set{Y})$,
  such that
  $X_1\in\Pi_R(Y_1)$
  and
  $X_2\in\Pi_R(Y_2)$
  with $Y_1 \neq Y_2 \in \set{Y}$,
  it holds
  \begin{align}
    X_1 \taudirect X_2 \ \ \mathrm{in} \ \mathcal{G}_\scm{L}
    \implies
    Y_1 \to Y_2 \ \ \mathrm{in} \ \mathcal{G}_\scm{H}.
  \end{align}\end{lemma}
\begin{proof}
    We report the proof in \Cref{proof:sufcon}
\end{proof}
\begin{corollary}[Sufficient Directed Paths]\label{lemma:corollarysufcon}
  Let $\scm{H}$ be a $\mat{T}$-abstraction of $\scm{L}$,
  where $\scm{H}$ and $\scm{L}$
  are two linear SCMs
  respectively on variables $\set{Y}$ and $\set{X}$ with graphs $\mathcal{G}_\scm{H}$ and $\mathcal{G}_\scm{L}$.
  Then, 
  for any pair
  of relevant variables $X_1,X_2\in\Pi_R(\set{Y})$,
  such that
  $X_1\in\Pi_R(Y_1)$
  and
  $X_2\in\Pi_R(Y_2)$
  with $Y_1 \neq Y_2 \in \set{Y}$,
  it holds that
  \begin{align}
    X_1 \anc X_2 \ \ \mathrm{in} \ \mathcal{G}_\scm{L}
    \implies
    Y_1 \anc Y_2 \ \ \mathrm{in} \ \mathcal{G}_\scm{H}.
  \end{align}
\end{corollary}
\begin{proof}
    We report the proof in \Cref{proof:corollarysufcon}
\end{proof}

In our previous results,
the faithfulness assumption
plays a fundamental role
to ensure that causal effect
is not canceled out
and thus propagates through $\mat{T}$-direct paths.
As we show in \Cref{ex:faithfulness},
whenever we allow for cancelling paths
we can construct a $\mat{T}$-abstraction
where two abstract variables
are not connected
despite the presence
of a $\mat{T}$-direct path
between their relevant variables.
\begin{example}[Unfaithful Concrete Model]\label{ex:faithfulness}
  Consider the following unfaithful linear SCM~$\scm{L}$
  where, given the weights as reported on the edges,
  the causal effect of $X_1$ on $X_4$ is canceled out.
  On the right,
  we show a linear SCM~$\scm{H}$. 
  \begin{center}
    \includegraphics[width=0.95\linewidth]{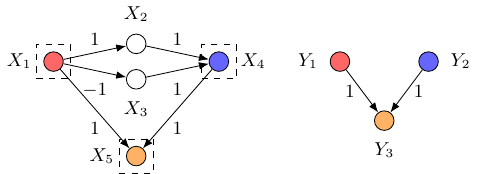}
  \end{center}
  Consider the linear abstraction function
  \begin{align}
    \mat{T} =
    \begin{bmatrix}
      1 & 0& 0& 0 & 0\\
      0 & 0& 0& 1 & 0\\
      0 & 0& 0& 0 & 1
    \end{bmatrix}^\tr
  \end{align}
  that maps each variable in $\set{X}$ in $\scm{L}$ to the corresponding variable in $\set{Y}$ in $\scm{H}$, e.g. the first column assigns $X_1$ to $Y_1$, the second column assigns $X_2$ to no high-level variable etc. We visualize the assignments by having the same color for the variables in the two models.
  Given this abstraction function,
  $\scm{H}$ is a $\mat{T}$-abstraction of $\scm{L}$,
  despite the $\mat{T}$-direct path $X_1 \taudirect X_4$
  between ${X_1\in\Pi_R(Y_1)}$ and ${X_4\in\Pi_R(Y_2)}$, not having a corresponding path $Y_1 \centernot\to Y_2$ (\Cref{proof:faithfulness}).
\end{example}

While the presence of a $\mat{T}$-direct path
between relevant variables
is a sufficient condition for the presence of an abstract edge,
the converse entails a stronger requirement.
It is in fact
necessary,
for an abstract edge ${Y_1 \to Y_2}$ to exist,
that for each variable in the relevant set of the source node $\Pi_R(Y_1)$
there exists a $\mat{T}$-direct path
to at least one relevant variable
of the target $\Pi_R(Y_2)$.
Intuitively,
any manipulation on a concrete variable
impacts its own abstract variable
and, consequently,
its descendants
in the abstract model.
To ensure consistency,
it is therefore necessary
that the manipulation
has an effect on the relevant variables
of the descendants (\Cref{ex:necessary}).
\begin{theorem}[Abstract Connectivity]\label{theo:connectivity}
  Let $\scm{H}$ be a $\mat{T}$-abstraction of $\scm{L}$,
  where $\scm{H}$ and $\scm{L}$
  are two linear SCMs
  respectively on variables $\set{Y}$ and $\set{X}$ with graphs $\mathcal{G}_\scm{H}$ and $\mathcal{G}_\scm{L}$.
  Then,
  there exists an edge ${Y_1 \to Y_2}$ in $\mathcal{G}_\scm{H}$
  if and only if
  for each $X_1\in\Pi_R(Y_1)$
  there exists $X_2\in\Pi_R(Y_2)$
  such that $X_1 \taudirect X_2$ in $\mathcal{G}_\scm{L}$.
\end{theorem}
\begin{proof}
    We report the proof in \Cref{proof:connectivity}
\end{proof}

By combining the sufficient condition
in \Cref{lemma:sufcon}
and the stronger necessary condition
in \Cref{theo:connectivity},
we can derive
a graphical condition
to show
whether
a model does \emph{not}
$\mat{T}$-abstract another
according to their graphs
and the set of relevant variables.
We formalize
this condition
in the following corollary,
which
we also describe
in \Cref{ex:necessary}.
\begin{corollary}[Connectivity Violation]\label{cor:necessary}
  Let $\scm{H}$ and $\scm{L}$
  be two linear SCMs
  respectively on variables $\set{Y}$ and $\set{X}$ with graphs $\mathcal{G}_\scm{H}$ and $\mathcal{G}_\scm{L}$. Consider a linear transformation~$\mat{T}$ between them leading to the sets of relevant variables $\Pi_R(\set{Y})$.
  If there exists three variables
  $X_1\in\Pi_R(Y_1)$,
  $X_2\in\Pi_R(Y_2)$, and
  $X_3\in\Pi_R(Y_1)$,
  such that
  both conditions hold
\begin{itemize}
    \item%
      $X_1 \taudirect X_2$
      in $\mathcal{G}_\scm{L}$, and
    \item%
      for any $X_4 \in \Pi_R(Y_2)$,
      $X_3 \centernot\taudirect X_4$
      is not in $\mathcal{G}_\scm{L}$,
\end{itemize}
    then $\scm{H}$ is not a $\mat{T}$-abstraction of $\scm{L}$.
\end{corollary}
\begin{proof}
    We report the proof in \Cref{proof:necessary}.
\end{proof}
\begin{example}[Abstract Connectivity Violation]\label{ex:necessary}
  Consider the following linear SCM~$\scm{L}$
  and a linear abstraction transformation~$\mat{T}$
  leading to the reported sets of relevant variables.
  \begin{center}
    \includegraphics[width=0.65\linewidth]{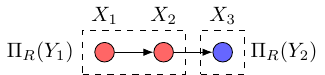}
  \end{center}
  Given the concrete edge $X_2 \taudirect X_3$,
  there must exist
  an abstract edge $Y_1 \to Y_2$ (\Cref{lemma:sufcon}).
  However,
  since for $X_3 \in \Pi_R(Y_1)$, the only path $X_1 \anc X_3$ to a variable in $\Pi_R(Y_2)$
  is mediated by the relevant variable $X_2$,
  it is not $\mat{T}$-direct
  and thus breaks the conditions of \Cref{theo:connectivity}, implying that there should be no abstract edge $Y_1 \to Y_2$ and leading to a contradiction.
  Intuitively,
  any two interventions
  ${i=(\Pi_R(Y_1) \gets [a, b])}$
  and
  ${i'=(\Pi_R(Y_1) \gets [a', b])}$
  where $a\neq a'$,
  have the sam ecausal effect on $X_3$,
  since there is not a $\mat{T}$-direct path $X_1 \taudirect X_3$.
  This breaks interventional consistency
  as $i,i'$ lead
  to distinct abstract interventions on $Y_1$
  and, thus, to different valus of $Y_2$.
  However, in the concrete model,
  the value of $X_3$ is the same
  regardless of $i,i'$ and, consequently,
  the value of $Y_2$ is constant.
  Therefore,
  given the portrayed graphs and relevant sets,
  for any choice of both structural
  and abstraction parameters
  any abstract model $\scm{H}$
  does not $\mat{T}$-abstract $\scm{L}$.
\end{example}

\subsection{Ordering of Concrete Blocks Induced by the Abstract Model}\label{subsec:varcnc}

Given our definition of $\mat{T}$-direct path,
we characterized the edges of the abstract graph
in terms of the connectivity of the relevant variables.
However,
despite not influencing directly
the abstraction function,
we can show that
irrelevant variables
still contribute to
abstract variables
and thus
have constraints
on their causal ordering.
In particular,
we identify
the set of concrete variables
whose corresponding exogenous variable
contributes to the noise term
of the abstract variable.
We call this subset
of variables
the \emph{concrete block} $\Pi(Y)$
of an abstract variable $Y$.
To define
the concrete block~$\Pi(Y)$,
we exploit the following corollary,
which proves that,
whenever the endogenous abstraction function
and the causal models are linear,
the exogenous abstraction function~$\gamma$
is necessarily a unique linear transformation.
\begin{corollary}[Exogenous Abstraction]\label{cor:exogabs}
  Let $\scm{H}=(\set{Y},\set{U},\set{g},\dist{P}_{\set{U}})$ be a {$\mat{T}$-abstraction} of 
  $\scm{L}=(\set{X},\set{E},\set{f},\dist{P}_{\set{E}})$,
  where $\scm{H}$ and $\scm{L}$
  are two linear SCMs.
  Then, the exogenous abstraction function
  $\gamma\colon\dom{\set{E}}\to\dom{\set{U}}$,
  has form
  \begin{align}
      \gamma(\vec{e}) = \mat{S}^\tr \vec{e},
  \end{align}
  where $\mat{S} = \mat{F} \mat{T}\mat{G}^\inv$
  and $\mat{F}, \mat{G}$
  are
  the linear
  transformations
  of
  respectively
  the reduced forms of $\scm{L}$ and $\scm{H}$, i.e., $\scm{L}(\vec{e}) = \mat{F}^T \vec{e}$ and $\scm{H}(\vec{u}) = \mat{G}^T \vec{u}$.
\end{corollary}
\begin{proof}
    We report the proof in \Cref{proof:exogabs}.
\end{proof}
\begin{definition}[Concrete Block]\label{def:block}
  Let $\scm{H}=(\set{Y},\set{U},\set{g},\dist{P}_{\set{U}})$ be a {$\mat{T}$-abstraction} of 
  $\scm{L}=(\set{X},\set{E},\set{f},\dist{P}_{\set{E}})$,
  where $\scm{H}$ and $\scm{L}$
  are two linear SCMs.
  We define
  the concrete block
  of each abstract variable $Y_j\in\set{Y}$
  as
  \begin{align}
    \Pi(Y_j) = \{ X_i \in\set{X} \mid s_{ij} \neq 0 \},
  \end{align}
  where $s_{ij}$
  is the $i$-th element
  on the $j$-th column
  of the matrix of the exogenous abstraction function $\mat{S}$.
  Moreover, we define the set of block variables $\Pi(\set{Y})$ of the abstract model~$\scm{H}$ as the union of the blocks of each $Y \in \set{Y}$. Formally,
  \begin{align}
   \Pi (\set{Y}) = \bigcup_{Y \in \set{Y}} \Pi(Y).
  \end{align} 
\end{definition}

We prove that the concrete block of an abstract variable contains the set of corresponding relevant variables.
In addition, it also contains all the irrelevant variables
that are
connected to one of these relevant variables
through a $\mat{T}$-direct path. %
\begin{lemma}[Block Composition]\label{lemma:blockordering}
  Let $\scm{H}$ be a $\mat{T}$-abstraction of $\scm{L}$,
  where $\scm{H}$ and $\scm{L}$
  are two linear SCMs
  respectively on variables $\set{Y}$ and $\set{X}$.
  Then,
  for any abstract variable $Y\in\set{Y}$,
  it holds
  $X\in\Pi(Y)$ if and only if
  \begin{itemize}
      \item $X\in\Pi_R(Y)$, or
      \item  $X \not \in \Pi_R(\set{Y})$, i.e., $X$ is irrelevant, and there exists $X'\in\Pi_R(Y)$ s.t. $X \taudirect X'$.
  \end{itemize}
\end{lemma}
\begin{proof}
    We report the proof in \Cref{proof:blockordering}.
\end{proof}

Intuitively, this result proves that
the irrelevant part of a block
lies between the relevant variables
of the abstract variable
and those of the abstract parents (\Cref{ex:block}).

\begin{example}[Block Composition]\label{ex:block}
  Given a concrete model $\scm{L}$ with six variables and an abstract model~$\scm{H}$ 
  with three variables
  such that $Y_1 \to Y_3 \gets Y_2$ that is a $\mat{T}$-abstraction of $\scm{L}$,
  we visualize
  a partition of concrete blocks induced by $\mat{T}$, where dashed lines denote sets of relevant variables.
  
  \begin{center}
    \includegraphics[width=0.6\linewidth]{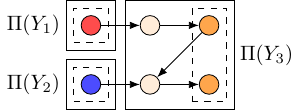}
\end{center}
Here block $\Pi(Y_3)$
  does not coincide
  with the set of relevant variables. The irrelevant variables in $\Pi(Y_3)$ have $\mat{T}$-direct paths to at least one of the relevant variables $\Pi_R(Y_3)$. 
\end{example}

In principle,
while sets of relevant variables are mutually disjoint,
the rest of the block could be shared
without breaking the consistency of the abstraction,
as we show in \Cref{ex:overlap}.

\begin{example}[Block Overlap]\label{ex:overlap}
  Let $\scm{L}$ be a linear SCM represented in the figure below,
  where the variable $X_2$ is in the block
  of both $Y_2$ and $Y_3$.
  Then, any abstract linear SCM $\scm{H}$
  that is a $\mat{T}$-abstraction of  $\scm{L}$ that induces these concrete blocks is not causally sufficient,
  since the exogenous terms $U_2, U_3$ in $\scm{H}$
  are a linear function of respectively
  $E_2, E_3$ and $E_2,E_4$ in $\scm{L}$,
  and hence they are not independent.
  Consequently,
  $Y_2$ and $Y_3$
  are confounded in any of these $\scm{H}$.
  \begin{center}
    \includegraphics[width=0.85\linewidth]{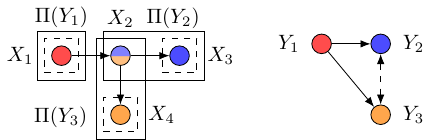}
  \end{center}
\end{example}

We prove that disjointness of irrelevant variables
in a block
is a necessary condition
to ensure abstract causal sufficiency.
\begin{lemma}[Disjoint Block]\label{lemma:disjointconstitutive}
  Let $\scm{H}$ be a $\mat{T}$-abstraction of $\scm{L}$,
  where $\scm{H}$ and $\scm{L}$
  are two linear SCMs
  respectively on variables $\set{Y}$ and $\set{X}$.
  If for any two distinct endogenous variables $Y_1, Y_2$
  it holds that $\Pi(Y_1) \cap \Pi(Y_2) \neq \emptyset$, then 
  the abstract model is not causally sufficient.
\end{lemma}
\begin{proof}
  We report the proof in \Cref{proof:disjointconstitutive}.
\end{proof}

We now prove our main result of this section, which shows
that the causal ordering
of the concrete blocks
must be consistent
with the abstract graph.
Intuitively, all relevant variables must follow
the abstract order (\Cref{theo:connectivity}) and
any irrelevant variable in a block
must precede
at least one relevant variable (\Cref{lemma:blockordering}).
Then,
given the causal sufficiency of the abstract model,
which ensures that blocks are disjoint (\Cref{lemma:disjointconstitutive}),
we can sort concrete blocks
according to the causal ordering of the abstract model. 
Further,
we can ignore
variables
that are not in any block
as they must
be last in the ordering
and thus
do not impact
abstract variables.
\begin{theorem}[Block Ordering]\label{theorem:absord}
  Let $\scm{H}$ be a $\mat{T}$-abstraction of $\scm{L}$,
  where $\scm{H}$ and $\scm{L}$
  are two linear SCMs
  respectively on variables $\set{Y}$ and $\set{X}$ with graphs $\mathcal{G}_\scm{H}$ and $\mathcal{G}_\scm{L}$.
  Then,
  for any valid topological ordering $\prec_{\scm{H}}$ of $\mathcal{G}_\scm{H}$
  there exists a valid ordering $\prec_{\scm{L}}$ of $\mathcal{G}_\scm{L}$
  such that
  for any $Y_1, Y_2, Y \in \set{Y}$:
  \begin{itemize}
      \item%
      $Y_1 \prec_{\scm{H}} Y_2 \iff
      \Pi(Y_1) \prec_{\scm{L}} \Pi(Y_2)$, and
      \item%
      $\Pi(Y) \prec_{\scm{L}} \big( \set{X} \setminus \Pi (\set{Y}) \big). %
        $
  \end{itemize}
\end{theorem}
\begin{proof}
    We report the proof in \Cref{proof:absord}.
\end{proof}

Given that
the ordering of concrete variables
depend on the abstract model,
we can show
that adding or removing
variables outside of
the blocks
still preserves $\mat{T}$-abstraction.

\begin{lemma}[Submodel Abstraction]\label{lemma:ignvar}
  Let $\scm{H}$ and $\scm{L}$
  be two linear SCMs
  respectively on variables $\set{Y}$ and $\set{X}$.
  Then,
  $\scm{H}$ is a $\mat{T}$-abstraction of $\scm{L}$
  if and only if
  $\scm{H}$ is a $\mat{T}$-abstraction of $\scm{L}^\prime$,
  where $\scm{L}^\prime$ is a submodel of $\scm{L}$
  defined on the subset of variables
  $\set{X}^\prime = \Pi(\set{Y})$, i.e., all of the variables in the concrete blocks.
\end{lemma}
\begin{proof}
    We report the proof in \Cref{proof:ignvar}.
\end{proof}

\subsection{Class of T-Concretizations of an Abstract Model}\label{subsec:weightconc}

After having characterized
the graphical structure
of two linear SCMs
in a $\mat{T}$-abstraction relation,
we now focus
on how abstraction
constraints the parameters
of the two models.
As we detailed in \Cref{lemma:ignvar},
variables that are not in any block
never cause,
either directly or indirectly,
relevant variables
and thus
can be ignored.
Therefore,
without loss of generality,
we consider only
variables
within the blocks $\Pi(\set{Y})$
of the abstract model.
Furthermore,
we permute
the weights of the concrete model
according to the abstract causal ordering with a permutation $\pi_{\scm{H}}$, derived from the valid ordering in \Cref{theorem:absord},
as in the following upper-diagonal block matrix
\begin{align}
  \mat{W} = \begin{bmatrix}
    \mat{W}_{11} & \mat{W}_{12} & \cdots & \mat{W}_{1b} \\
    \mat{0} & \mat{W}_{22} & \cdots & \mat{W}_{2b} \\
    \vdots & \vdots & \ddots & \vdots \\
    \mat{0} & \mat{0} & \cdots & \mat{W}_{bb}
  \end{bmatrix},
\end{align}
where
we denote
by $\mat{W}_{hk}\in\real^{N_h \times N_k}$
the submatrix
containing the edges
from the concrete block $\Pi(Y_h)$
to $\Pi(Y_k)$.
Under the same permutation $\pi_{\scm{H}}$,
we can also block-wise define
the linear abstraction transformation as follows
\begin{align}
  \mat{T} = \begin{bmatrix}
    \vec{t}_1 & \mat{0} & \cdots & \mat{0} \\
    \mat{0} & \vec{t}_2 & \cdots & \mat{0} \\
    \vdots & \vdots & \ddots & \vdots \\
    \mat{0} & \mat{0} & \cdots & \vec{t}_b.
  \end{bmatrix}
\end{align}
where each $\vec{t}_k$ is a vector of size $N_k$.
Each of these vectors can still have zero entries for the irrelevant variables.
Notably,
due to the fact that
no irrelevant variable
follows a relevant one
in the same block,
the last component
of each vector is non-zero.

Given the same permutation $\pi_{\scm{H}}$,
the exogenous transformation~$\mat{S}$
necessarily follows the same structure
and is defined by
the endogenous abstraction function
and the causal relations among variables in the same block.
As a direct consequence,
the exogenous
and the endogenous
transformations
coincide whenever
a block
lacks internal causal relations
and, consequently,
all variables in the block
are relevant.
\begin{lemma}[Exogenous Abstraction]\label{lemma:exoabs}
  Let $\scm{H}=(\set{Y}, \set{U}, \mat{M},  \dist{P}_{\set{U}})$
  and $\scm{L}=(\set{X},  \set{E}, \mat{W}, \dist{P}_{\set{E}})$
  be two linear SCMs
  such that $\scm{H}$
  is a $\mat{T}$-abstraction
  of $\scm{L}$, such that $\mat{W}$ follows permutation $\pi_{\scm{H}}$.
  Then, the exogenous abstraction function $\gamma\colon\dom{\set{E}}\to\dom{\set{U}}$
  is unique and
  has form ${\gamma(\vec{e}) = \mat{S}^\tr \vec{e}}$
  for a linear transformation ${\mat{S}\in\real^{d \times b}}$
  defined as the upper-diagonal block matrix
  \begin{align}
    \mat{S} = \begin{bmatrix}
      \vec{s}_1 & \mat{0} & \cdots & \mat{0} \\
      \mat{0} & \vec{s}_2 & \cdots & \mat{0} \\
      \vdots & \vdots & \ddots & \vdots \\
      \mat{0} & \mat{0} & \cdots & \vec{s}_b,
    \end{bmatrix}
  \end{align}
  where
  $\vec{s}_k = \mat{F}_{kk} \vec{t}_k = {(\mat{I} - \mat{W}_{kk})}^\inv\vec{t}_k$
  for any $Y_k\in\set{Y}$.%
\end{lemma}
\begin{proof}
    We report the proof in \Cref{proof:exoabs}.
\end{proof}

Given the structure
and the ordering induced
by the abstraction function,
we introduce a provably
equivalent formulation
of $\mat{T}$-abstraction
entirely based
on the model parameters.
In this way,
we guarantee interventional consistency
on all possible abstract hard interventions
as a property of the weights of the two linear SCMs.
Further,
by assessing abstraction in closed-form,
we can
characterize the set of $\mat{T}$-concretizations
of an abstract model
(\Cref{ex:multiabs}).
\begin{theorem}[Block Abstraction]\label{theo:concretization}
  Let $\scm{H}=(\set{Y}, \set{U}, \mat{M}, \dist{P}_{\set{U}})$
  and $\scm{L}=(\set{X}, \set{E}, \mat{W}, \dist{P}_{\set{E}})$
  be two linear SCMs with graphs $\mathcal{G}_{\scm{H}}$ and $\mathcal{G}_{\scm{L}}$ respectively.
  Then $\scm{H}$ is a linear $\mat{T}$-abstraction of $\scm{L}$
  if and only if
  for any valid topological ordering $\prec_{\scm{H}}$ of $\mathcal{G}_{\scm{H}}$
  there exists a valid ordering $\prec_{\scm{L}}$ of $\mathcal{G}_{\scm{L}}$
  such that,
  for any $Y_i,Y_j\in\set{Y}$ it holds
  \begin{align}\label{eq:weightconsistency}
    Y_i \prec_{\scm{H}} Y_j &\iff \Pi(Y_i) \prec_{\scm{L}} \Pi(Y_j), \ \mathrm{and}\\
    \mat{W}_{ij}\vec{s}_j &= m_{ij} \vec{t}_i,
  \end{align}
  where $\mat{W}_{ij}$ is the $i$-th element on the $j$-th column of $\mat{W}$, and $m_{ij}$ is the $i$-th element on the $j$-th column of $\mat{M}$.
\end{theorem}
\begin{proof}
  We report the proof in \Cref{proof:concretization}.
\end{proof}
\begin{example}[$\mat{T}$-Concretization Class]\label{ex:multiabs}
  Let $\scm{H}$ be an abstract causal model
  with two variables such that
  ${Y_1 \to Y_2}$ with unitary weight,
  and let $\mat{T}$
  be
  the following transformation%
  \begin{align}
    \mat{T} =
    \begin{bmatrix}
      1 & 1& 0& 0\\
      0 & 0& 1& 1
    \end{bmatrix}^\tr,
  \end{align}
  Then,
  of the three following linear SCMs,
  we can easily verify
  that
  only the first two models
  are $\mat{T}$-abstracted
  by $\scm{H}$.
  \begin{center}
    \includegraphics[width=0.30\linewidth]{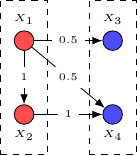}
    \hfill
    \includegraphics[width=0.30\linewidth]{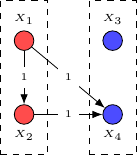}
    \hfill
    \includegraphics[width=0.30\linewidth]{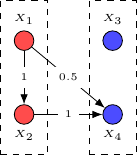}
  \end{center}
  Given the identical inner-block connections,
  the exogenous abstraction function
  is the same for all three models, as in
  \begin{align}
    \vec{s}_1 = \begin{bmatrix}
      2\\1
    \end{bmatrix},
    \quad
    \vec{s}_2 = \begin{bmatrix}
      1\\1
    \end{bmatrix}.
  \end{align}
  Then, only for the first two models
  it holds $\mat{W}_{12}\vec{s}_2 = \vec{t}_1$.
\end{example}

\begin{algorithm}[t]
  \SetAlgoLined%
  \KwIn{%
    Abstract adjacency matrix $\mat{M}\in\real^{b\times b}$
  Abstraction function $\mat{T}\in\real^{d \times b}$}
  \KwResult{%
    Concrete adjacency matrix $\mat{W}\in\real^{d\times d}$
  }
  \SetCommentSty{small}
  \SetKwComment{Comment}{$\triangleright$ }{}
  $\mat{W} \gets \mat{0}$
  \Comment*[f]{Init Concrete Weights}\\
  \For(\Comment*[f]{Abstract Target Node}){$Y_j\in\set{Y}$}{
    $N_j \gets |\Pi(Y_j)|$\\
    $\mat{W}_{jj} \gets \operatorname{RandomDAG}(N_j)$
    \Comment*[f]{Target Block Weights}\\
    $\vec{s}_j \gets {(\mat{I} - \mat{W}_{jj})}^\inv \vec{t}_j$\\
    \For(\Comment*[f]{Abstract Source Node}){$Y_i\in\set{Y}$}{
      \For(\Comment*[f]{Source Block}){$X_k\in\Pi(Y_i)$}{
        $\vec{v} \sim \{\vec{v}\in\real^{N_j} \mid \sum_{h=1}^{N_j} v_h = 1\}$\\
        {$\vec{c} \gets \vec{v} / \vec{s}_j$
        \Comment*[f]{Right-Inverse of $\vec{s}_j$}}
        ${[\mat{W}_{ij}]}_{k, \colon} \gets m_{ij}{[\vec{t}_i]}_k \vec{c}^\tr$
        \Comment*[f]{Assign $k$-th row}
      }
    }
  }
  \caption{%
    $\mat{T}$-Concretization Sampling
  }%
  \label{alg:samplingblocks}
\end{algorithm}

By building on our novel formulation,
we define a complete and sound procedure
to sample concrete models
from an abstract adjacency matrix
and a linear abstraction function (\Cref{alg:samplingblocks}).
First,
for each abstract target variable $Y_j$,
the algorithm samples
the inner-block weights~$\mat{W}_{jj}$,
where we assume
that any irrelevant variable
has at least a relevant variable as a descendant.
Consequently,
all variables
are members of the block.
Then,
for each source variable~$Y_i$,
the algorithm samples consistent coefficients~$\mat{W}_{ij}$
respecting \Cref{theo:concretization} by first sampling
a right-inverses
of the exogenous abstraction function~$\vec{s}_j$.
Since the generated model
follows
the abstract causal ordering
and \Cref{theo:concretization}
by construction,
it is a valid concretization.

\section{Abstract Information for Non-Gaussian Linear Discovery}\label{sec:method}

\begin{algorithm}[t]

  \SetAlgoLined%
  \KwIn{%
    Concrete Observational Dataset $\dset_{\scm{L}}$,\\
    \hspace{3em}Joint Observational Dataset $\dset_J$.
  }
  \KwResult{%
    Abstraction function $\hat{\mat{T}}\in\real^{d \times b}$,\\
    \hspace{3.3em}Abstract adjacency matrix $\hat{\mat{M}}\in\real^{b \times b}$,\\
    \hspace{3.3em}Concrete adjacency matrix $\hat{\mat{W}}\in\real^{d \times d}$.
  }
  \SetCommentSty{small}
  \SetKwComment{Comment}{$\triangleright$ }{}
  $\hat{\mat{T}} \gets
  \argmin_{\mat{T}\in\real^{b \times d}}
  \sum_{(\vec{x}, \vec{y}) \in \dset_J} {
    {\| \vec{x}^\tr \mat{T} - \vec{y}^\tr \|}^2_2
  }$\;
  \For(\Comment*[f]{Select Relevant Variables}){$Y_i \in \set{Y}$}{
    $\hat{\Pi}_R(Y_i) \gets \{ X_k \in \set{X}\mid {[\hat{\vec{t}_i}]}_k \neq 0 \}$
  }
  $\dset_{\hat{\scm{H}}} \gets \{\hat{\mat{T}}^\tr \vec{x}  \mid \vec{x} \in \dset_{\scm{L}} \}$
  \Comment*[f]{Create Abstract Dataset}\\
  $\hat{\mat{M}} \gets \text{DirectLiNGAM}(\dset_{\hat{\scm{H}}}, \emptyset)$
  \Comment*[f]{Abstract Discovery}\\
  $\set{K} \gets \emptyset$\\
  \For(\Comment*[f]{Collect Prior Knowledge}){$Y_i, Y_j \in \set{Y}$}{
    \If(\Comment*[f]{Check Ancestorship in $\hat{\mat{M}}$}){$Y_i \centernot\anc Y_j$}{
      \For{$X_k \in \hat\Pi_R(Y_i)$, $X_h \in \hat\Pi_R(Y_j)$}{
          $\set{K} \gets \set{K}
          \cup \{ X_k \centernot\anc X_h \}$
      }
    }
  }
  $\hat{\mat{W}} \gets \text{DirectLiNGAM}(\dset_{\scm{L}}, \set{K})$
  \Comment*[f]{Concrete Discovery}\\
  \caption{Abs-LiNGAM}%
  \label{alg:abslingam}
\end{algorithm}

In this section, we introduce Abs-LiNGAM (\Cref{alg:abslingam}),
a strategy to exploit our results
on $\mat{T}$-abstraction
to speedup observational causal discovery
of linear non-Gaussian models, e.g. LiNGAM~\citep{shimizu2011directlingam}.
The intuition is that 
whenever we have a $\mat{T}$-abstraction
of an unknown model to learn,
we can exclude all the candidate solutions
not satisfying the graphical conditions we presented in the previous sections.
Furthermore,
in Abs-LiNGAM,
we demonstrate
how to infer
prior knowledge for the concrete model
from a small number
of paired concrete-abstract samples,
even when the abstract model
and the abstraction function
are unknown, and an abstract dataset is not directly available.
In the following,
we formalize the data-generation process
and the steps of Abs-LiNGAM\@.

\subsection{Data-Generation Process}\label{subsec:dgp}

As in many real-world applications,
where observations are produced
by sensors or other data-collecting devices,
we assume that samples
from the low-level concrete model
have a significantly larger availability
than high-level abstract samples.
We formalize this
intuition
by defining two datasets
\begin{align}
  \dset_{\scm{L}} &\sim \dist{P}_{\scm{L}}\\
  \dset_J &\sim \dist{P}_{\scm{L}, \scm{H}},
\end{align}
where the former contains
concrete samples only
and the latter
paired observations
from the joint observational distribution
of both models,
such that $|\dset_{J}| \ll |\dset_{\scm{L}}|$.
Therefore,
we define the following
data-generating process,
where we produce
a significantly lower number
of abstract samples.
\begin{align}
  \vec{e}^{(i)} &\sim \operatorname{Exponential}
                &\text{for } i=1,\dots,|\dset_{\scm{L}}|,\\
  \vec{x}^{(i)} &= \scm{L}(\vec{e}^{(i)})
                &\text{for } i=1,\dots,|\dset_{\scm{L}}|,\\
  \vec{y}^{(i)} &= \scm{H}(\gamma(\vec{e}^{(i)}))
                &\text{for } i=1,\dots,|\dset_{J}|.
\end{align}
Since we assume linear and non-Gaussian data,
the models are identifiable
in the limit of infinite data~\citep{shimizu2006linear}.
In \Cref{app:noisy},
we discuss preliminary results
to tackle an additional
scenario where
we consider
abstract observations
to be perturbed
by random noise.

\subsection{Abs-LiNGAM}

\paragraph{T-Reconstruction.}
Since we assume a linear transformation,
we can fit the abstraction function
from the joint dataset~$\dset_J$
by solving a least-squares problem~\citep{trefethen2022numerical}.
Then,
for each abstract variable~$Y_i\in\set{Y}$,
we can identify
its set of relevant variables~$\hat{\Pi}_R(Y)$,
as 
\begin{align}
  \hat{\Pi}_R(Y_i) = \{X_k \mid {[\hat{\vec{t}}_i]}_k \neq 0\}.
\end{align}
In practice,
we mask the coefficients
of the fitted abstraction
transformation~$\hat{\mat{T}}$
with a small threshold
to handle numerical instability,
which, whenever a sufficient number
of joint samples $|\mathcal{D_J}|$ is available,
ensures that each relevant block pertains to a single abstract variable.

\paragraph{Abstract Causal Discovery.}
Then, we focus on learning
the abstract causal structure
from data.
Since
we assume
abstract samples
to be scarce,
even in our simplified setting of linear and non-Gaussian models,
the abstract model
might not be discoverable
by the high-level samples
in the joint dataset~$\dset_J$ alone.
However,
after having identified the abstraction function,
we can use it on the concrete dataset
to abstract each sample as in
\begin{align}
  \dset_{\hat{\scm{H}}}= \{ \hat{\mat{T}}^\tr \vec{x} \mid \vec{x} \in \dset_{\scm{L}} \}.
\end{align}
In fact,
whenever
the target model
is a $\mat{T}$-abstraction,
the observational consistency property
ensures that
abstracting concrete samples
is equivalent to directly sampling
from the abstract distribution,
as in the data-generating process.
Then,
we can use the newly generated
abstract samples
with any causal discovery algorithm
for linear non-Gaussian models.

\paragraph{Concrete Causal Discovery}
Finally,
we can use the constraints induced by the abstract model to speedup discovery
of the concrete causal model.
As an immediate consequence of \Cref{theo:connectivity},
the existence
of an abstract directed path
$Y_i \anc Y_j$
entails the existence
of at least
a concrete directed path
between variables
in the corresponding relevant blocks $\Pi_R(Y_i)$ and $\Pi_R(Y_j)$.
We cannot,
however,
directly
infer which
of the possibly many
ancestral relations
the concrete model contains.
On the other hand,
whenever an abstract path does \emph{not} exist,
we can infer that
any variable
in the source block
does not cause,
neither directly or indirectly,
any variable in the target block.
We can therefore
restrict the search space
of the concrete causal discovery problem
by excluding
all solutions
that do not satisfy
the following set of constraints
\begin{equation}
  \begin{aligned}
  \set{K} = \{
    &X_k \centernot\anc X_h
    \mid
    \,X_k \in \Pi_R(Y_i)\\
    \land
    &\,X_h \in \Pi_R(Y_j)
    \land
    \,Y_i \centernot\anc Y_j
  \}.
  \end{aligned}
\end{equation}
We use the DirectLiNGAM algorithm~\citep{shimizu2011directlingam}
to solve the concrete causal discovery problem,
as it can integrate
prior knowledge
in the form of
forbidden direct paths
and thus restrict the set of candidate solutions.

\definecolor{darkgreen}{rgb}{0.0, 0.5, 0.0}
\definecolor{darkorange}{rgb}{0.8, 0.4, 0.0}
\begin{figure*}
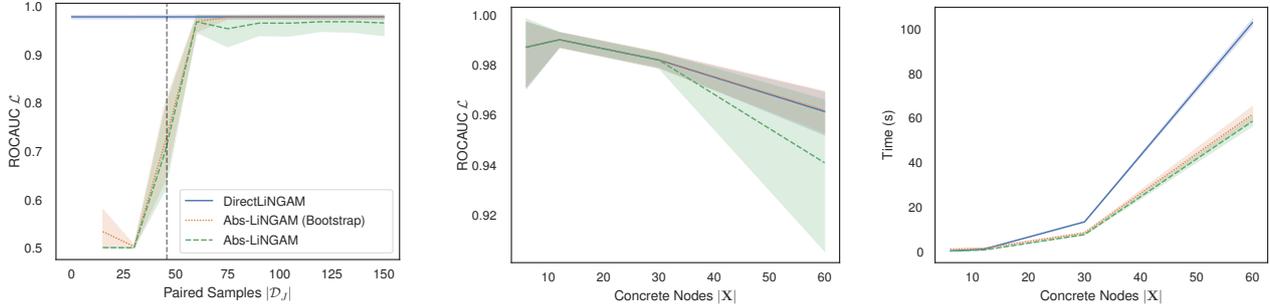

  \centering
  \begin{subfigure}[b]{0.33\textwidth}
    \centering
    \resizebox{\textwidth}{!}{%
      \input{figures/exp1_concrete_roc_auc_small.pgf}
    }
    \caption{%
      Performance over Paired Samples~$|\dset_J|$
    }\label{fig:expA}
  \end{subfigure}
  \hfill
  \begin{subfigure}[b]{0.66\textwidth}
    \centering
    \resizebox{0.49\textwidth}{!}{%
      \input{figures/exp3_concrete_roc_auc_small.pgf}
    }
    \resizebox{0.49\textwidth}{!}{%
      \input{figures/exp3_time_small.pgf}
    }
    \caption{%
      Performance
      and Execution Time (s)
      over Concrete Graph Size~$|\set{X}|$
    }\label{fig:expB}
  \end{subfigure}
  \caption{%
    We report the performance
    of Abs-LiNGAM
    for (a) an increasing number
    of paired samples~$|\dset_J|$~%
    and (b) an increasing number
    of concrete nodes~$|\set{X}|$~.%
    We plot a variant
    of Abs-LiNGAM
    where we bootstrap
    the abstract causal discovery step
    with five repetitions.
    We report the area under the ROC curve
    and the execution time
    over 30 runs
    on randomly generated
    Erd\H{o}s-R\'enyi abstract graphs
    with $b=5$ nodes
    and 8 edges.
    In the first experiment,
    we sample for each abstract graph
    a concrete model
    with random size $|\set{X}| \in [25, 50]$.
    In the second experiment,
    we also vary the number of paired samples
    to always be twice the number of concrete nodes.
  }\label{fig:experiments}
\end{figure*}

\section{Experimental Results}\label{sec:experiments}

In this section,
we discuss
our analysis
on the performance
of Abs-LiNGAM (\Cref{alg:abslingam})
on simulated data.
In particular,
we validate
whether
a small amount
of paired concrete-abstract
observations
can reduce
the search space,
and thus the execution time,
of DirectLiNGAM~\citep{shimizu2011directlingam},
without compromising
the quality of the retrieved
concrete causal structure.
As baseline,
we compare
against applying DirectLiNGAM
to the concrete dataset
without any abstract-induced
prior knowledge.

For each run,
we sample 
the parameters
of an abstraction function
and
of an abstract linear SCM\@.
We then generate
a concrete causal model
by sampling
one of the
possible
$\mat{T}$-concretizations
of the abstract model with \Cref{alg:samplingblocks}.
We provide
details on our experimental setup
and additional results
respectively
in \Cref{app:dataset} and in
\Cref{app:additional}.

We study
the performance
of Abs-LiNGAM
for an increasing number
of paired samples~(\Cref{fig:expA}).
Since Abs-LiNGAM is a multi-step algorithm,
the quality of the retrieved
concrete causal model
strictly depends on the correctness
of the abstraction function, the consequent generated abstract data and abstract causal discovery.
As expected,
whenever
the size of the paired dataset~$|\dset_J|$
is too small,
Abs-LiNGAM
wrongly identifies
concrete paths
as forbidden
and, compared to the baseline,
fails to retrieve the correct
concrete causal model.
However,
whenever the number
of paired samples
approaches the number
of concrete nodes~$|\set{X}|$,
Abs-LiNGAM
performs similarly
to the baseline
and correctly retrieves
the concrete causal model.
We observe
the same trend
for concrete graphs
of increasing size~(\Cref{fig:expB}),
highlighting
how 
prior knowledge induced from the abstract model
significantly reduces
the execution time
compared to the baseline.

Furthermore,
we found that
bootstrapping abstract causal discovery, i.e.,
aggregating several iterations
on randomly extracted sub-datasets,
improves
the performance
on the downstream
concrete discovery task
without noticeably affecting
the execution time,
which is still dominated
by the final concrete causal discovery run.

\section{Related Works}

Several works addressed the problem
of clustering together variables
to reduce dimensionality
and maintain the
identifiability of causal effect.
Both \citet{anand2023causal} and \citet{wahl2023foundations}
deal with the problem of partitioning
a causal graph into clusters
where causal relations at the micro-level
are translated 
as causal edges at the macro-level.
\citet{tikka2023clustering} 
study instead a particular class of groups,
which they define as \emph{transit clusters},
where only part of the variables
are allowed to have
ingoing or outgoing edgs
from the cluster.

Differently from previous works,
our work
focuses instead
on the necessary conditions
for causal abstraction
and results in different
definitions for the grouping of micro-variables.
It is however
an interesting direction
to assess
whether different assumptions,
for instance on the intervention map,
might lead to comparable definitions.

In parallel, several recent papers explored the problem of fitting an abstraction function from data by focusing on either discrete~\citep{zennaro2023jointly,felekis2024causal} or linear~\citep{kekic2023targeted,geiger2024finding} SCMs. Notably, apart from interventional samples, all these works assume to have at least partial knowledge of the graphs, the intervention map, or the set of concrete relevant variables corresponding to each abstract one.

Based on our theoretical results on the graphical and parametric conditions of linear abstraction for linear causal models,
we instead propose 
to learn both the abstract and the concrete model,
and their abstraction function directly from observational data and
without any prior knowledge or any constraint on the graphical structure of the two models.

\section{Conclusion}

In this paper,
we studied
the necessary and sufficient conditions
on the causal ordering
and the parameters
for two linear Structural Causal Models
to be in a linear abstraction.
Furthermore,
we introduced
the first procedure
to sample from
the set of all possible
concretizations of an abstract SCM, which can be used in other abstraction applications\@.

We also proposed Abs-LiNGAM,
a strategy
to speedup causal discovery
of a linear non-Gaussian concrete causal model
given an additional dataset
of paired observations
on concrete and abstract variables.
Finally,
we empirically
highlighted how Abs-LiNGAM
leverages
abstract information
to reduce the search space
and improve
execution time
without sacrificing
on the
quality
of the discovered
structure.

 An interesting direction for future work
 is to extend our results
 to non-linear models and non-linear abstraction functions,
 and to tackle the causal sufficiency assumption,
 which requires full-observability of the concrete realizations.

\begin{acknowledgements}
This work has been supported by EU-EIC EMERGE (Grant
No. 101070918), by H2020 TAILOR (Grant
No. 952215) and by the EU NextGenerationEU programme
under the funding schemes PNRR-PE-AI (PE00000013)
FAIR --- Future Artificial Intelligence Research.
\end{acknowledgements}

\clearpage
\appendix
\section*{Appendix}

We organize the Appendix as follows.
In \Cref{app:glossary},
we report further information
on our notation
by summarizing it in a glossary.
Then, in \Cref{app:proofs},
we report all the proof
for the theoretical results
discussed in the main body.
Finally, in \Cref{app:dataset},
we present further details
on the generative process
of the synthetic datasets
used for our empirical study,
of which we report additional results
in \Cref{app:additional}.

\section{Glossary}\label{app:glossary}

\begin{tabular}{l l}
  \textbf{Notation} & \textbf{Definition} \\
  \toprule
  $\set{X}$ & Set of endogenous concrete variables \\
  $\set{E}$ & Set of exogenous concrete variables \\
  $\set{Y}$ & Set of endogenous abstract variables \\
  $\set{U}$ & Set of exogenous abstract variables \\
  $d$ & Number of concrete variables \\
  $b$ & Number of abstract variables \\
  $\scm{L}$ & Concrete Causal Model \\
  $\scm{H}$ & Abstract Causal Model \\
  $\mat{W}$ & Weighted Adjacencies of $\scm{L}$ \\
  $\mat{M}$ & Weighted Adjacencies of $\scm{H}$ \\
  $\mat{F}$ & Reduced Form of $\scm{L}$ \\
  $\mat{G}$ & Reduced Form of $\scm{H}$ \\
  $\tau$ & Endogenous Abstraction Function \\
  $\gamma$ & Exogenous Abstraction Function \\
  $\mat{T}$ & Linear Endog. Abstraction Transformation \\
  $\mat{S}$ & Linear Exog. Abstraction Transformation \\
  $\vec{t}_j$ & Vector in $\mat{T}$ abstracting $Y_j$ from $\Pi_R(Y_j)$ \\
  $\vec{s}_j$ & Vector in $\mat{S}$ abstracting $U_j$ from $\vec{e}_{\Pi(Y_j)}$\\
  $\Pi_R(Y)$ & Set of relevant variables for $Y$ \\
  $\Pi(Y)$ & Block of $Y$ \\
  $N_i$ & Number of variables in $\Pi(Y_i)$ \\
  $\mat{W}_{ij}$ & Submatrix of weights from $\Pi(Y_i)$ to $\Pi(Y_j)$ \\
  $\mat{F}_{ij}$ & Submatrix of sub-model from $\Pi(Y_i)$ to $\Pi(Y_j)$ \\
  $\dset_{\scm{L}}$ & Dataset sampled from $\dist{P}_{\set{X}}$ \\
  $\dset_J$ & Dataset sampled from the joint $\dist{P}_{\set{X},\set{Y}}$ \\
\end{tabular}

\section{Proofs}\label{app:proofs}

\subsection{Lemma~\ref{lemma:disjointrelevant}}\label{proof:disjointrelevant}
\paragraph{\Cref{lemma:disjointrelevant} (Disjoint Relevant).}
  Let $\scm{H}$ be a $\mat{T}$-abstraction of $\scm{L}$,
  where $\scm{H}$ and $\scm{L}$
  are two linear SCMs
  respectively on variables $\set{Y}$ and $\set{X}$.
  Then,
  for any pair of distinct abstract variables $Y_1, Y_2\in\set{Y}$,
  it holds that $\Pi_R(Y_1)\cap\Pi_R(Y_2) = \emptyset$,
  where $\Pi_R(Y_1) \neq \emptyset$ and $\Pi_R(Y_2) \neq \emptyset$.
\begin{proof}
  Firstly,
  we show that
  given an abstract intervention ${j = (Y_1 \gets k)}$
  on $Y_1$,
  any concrete intervention $i$
  such that $\omega(i) = j$
  must fix all relevant variables~$\rel{Y_1}$.
  Otherwise,
  if we assume the existence
  of a non-intervened variable $X_s\in\rel{Y_1}$
  the function $\scm{L}^i_{\rel{Y_1}}$
  would be non-constant.
  Therefore,
  since
  $\tau_{Y_1}$ depends on $X_s$
  by definition of relevant variable,
  interventional consistency would not hold,
  as in
  \begin{align}
    \tau_{Y_1} \circ \scm{L}^i_{\Pi_R(Y_1)}     
    \neq
    \scm{H}^j_{Y_1} \circ \gamma
    = k.
  \end{align}
  Therefore,
  for any abstract intervention~$j=(Y_1 \gets k)$,
  the corresponding concrete interventions
  must have form
  \begin{align}
    i = (\Pi_R(Y_1) \gets \vec{v}),
  \end{align}
  for any vector $\vec{v}$ such that $\tau_{Y_1}(\vec{v}) = k$,
  without intervening on further relevant variables.

  We firstly
  prove \Cref{lemma:disjointrelevant}
  whenever $Y_1 \centernot\anc Y_2$.
  Then,
  we assume the existence
  of a non-empty subset $\set{V} = \Pi_R(Y_1) \cap \Pi_R(Y_2)$
  of shared variables.
  Since $Y_1$ has no causal effect on $Y_2$,
  given an high-level intervention
  ${j=(Y_1 \gets k)}$,
  it must hold that
  \begin{align}
      \scm{H}^j_{Y_2} = \scm{H}_{Y_2}.
  \end{align}
  However, by intervening on $Y_1$,
  any concretization must also fix $\set{V}$.
  Therefore, we prove the property by contradiction, as
  \begin{align}
    \scm{H}^j_{Y_2} \circ \gamma &=
    \tau_{Y_2} \circ \scm{L}^i_{\Pi_R(Y_2)} \\
    & \neq \tau_{Y_2} \circ \scm{L}_{\Pi_R(Y_2)} \\
    &= \scm{H}_{Y_2} \circ \gamma\\
    &\implies 
    \scm{H}^j_{Y_2} \neq \scm{H}_{Y_2},
  \end{align}
  given the surjectivity of $\gamma$
  and the lack of cancelling paths.

  Finally, we can tackle the last scenario,
  where $Y_1 \anc Y_2$,
  by showing that $\Pi_R(Y) \cap \Pi_R(\an{Y}) = \emptyset$,
  where $\an{Y}$ is the set of ancestors of $Y$.
  Given the model acyclicity,
  for any abstract intervention~$j=(Y\gets k)$,
  it must hold
  \begin{align}
      \scm{H}^j_{\an{Y}} = \scm{H}_{\an{Y}}.
  \end{align}
  However,
  if the relevant variables of $Y$
  were to overlap
  with the relevant variables of its ancestors,
  we could show that
  \begin{align}
    \scm{H}^j_{\an{Y}} \circ \gamma
    &= \tau_{\an{Y}} \circ \scm{L}^i_{\an{Y}} \\
    &\neq \tau_{\an{Y}} \circ \scm{L}_{\an{Y}} \\
    &= \scm{H}_{\an{Y}} \circ \gamma\\
    &\implies 
    \scm{H}^j_{\an{Y}} \neq \scm{H}_{\an{Y}}.
  \end{align}
  Therefore,
  since interventional consistency does not hold,
  $\scm{H}$ is not a $\mat{T}$-abstraction of $\scm{L}$,
  which contradicts the hypothesis
  and concludes the proof.
\end{proof}

\subsection{Corollary~\ref{cor:constructive}}\label{proof:constructive}
\paragraph{\Cref{cor:constructive} (Constructive Abstraction).}
    Let $\scm{H}$
    be a strong $\tau$-abstraction
    of $\scm{L}$ where
    $\scm{H}$ and $\scm{L}$ are linear SCMs
    and
    $\tau$ is a linear transformation.
    Then,
    $\scm{H}$ is a constructive $\tau$-abstraction
    of $\scm{L}$.
\begin{proof}
    By definition of linear transformation,
    the set of low-level variables
    on which an abstract variable~$Y\in\set{Y}$
    depends through the linear abstraction function~$\tau$
    coincides with its
    set of relevant variables~$\rel{Y}\subseteq\set{X}$.
    Therefore,
    by showing
    that the relevant sets are disjoint
    whenever the SCMs $\scm{H}$ and $\scm{L}$
    are linear,
    a $\mat{T}$-abstraction on linear SCMs
    is also a constructive abstraction.
    By definition of $\mat{T}$-abstraction,
    this is equivalent
    to state that
    a linear $\tau$-abstraction
    on linear SCMs
    is a constructive abstraction
    under our assumption
    on the absence of cancelling paths.
\end{proof}

\subsection{Lemma~\ref{lemma:sufcon}}\label{proof:sufcon}
\paragraph{\Cref{lemma:sufcon} (Sufficient Directed Paths)}
  Let $\scm{H}$ be a $\mat{T}$-abstraction of $\scm{L}$,
  where $\scm{H}$ and $\scm{L}$
  are two linear SCMs
  respectively on variables $\set{Y}$ and $\set{X}$ with graphs $\mathcal{G}_\scm{H}$ and $\mathcal{G}_\scm{L}$.
  Then, 
  for any pair
  of relevant variables $X_1,X_2\in\Pi_R(\set{Y})$,
  such that
  $X_1\in\Pi_R(Y_1)$
  and
  $X_2\in\Pi_R(Y_2)$
  with $Y_1 \neq Y_2 \in \set{Y}$,
  it holds
  \begin{align}
    X_1 \taudirect X_2 \ \ \mathrm{in} \ \mathcal{G}_\scm{L}
    \implies
    Y_1 \to Y_2 \ \ \mathrm{in} \ \mathcal{G}_\scm{H}.
  \end{align}
\begin{proof}
  Let $Y_1, Y_2$ be two distinct abstract variables
  and
  let $i,i'$
  be two concrete interventions
  that fix any \emph{relevant} variable
  except for those in the relevant set $\Pi_R(Y_2)$,
  and
  whose assignments differ only in $X_1\in\Pi_R(Y_1)$.
  Formally,
  \begin{align}
  i &= (\set{V} \gets \vec{v},\,\Pi_R(Y_1) \gets \vec{c})\\
  i' &= (\set{V} \gets \vec{v},\,\Pi_R(Y_1) \gets \vec{c}'),
  \end{align}
  where
  \begin{align}
      \set{V}
      &=
      \bigcup_{Y \in \set{Y} \setminus \{Y_1, Y_2\}} \Pi_R(Y).
  \end{align}
  Given $X_1 \taudirect X_2$,
  there exists at least a directed path
  composed only of non-relevant variables,
  that are therefore non-intervened.
  Consequently,
  due to the \emph{faithfulness} assumption,
  the concrete model does not have cancelling paths
  and, therefore,
  an intervention on a variable
  always has an effect on its descendants.
  In particular,
  since $i,i'$ constrain $X_1$
  to two different values,
  it holds that
  \begin{align}
      \scm{L}^i_{X_2} &\neq \scm{L}^{i'}_{X_2}\\
      \tau_{Y_2}\circ\scm{L}^i &\neq \tau_{Y_2}\circ\scm{L}^{i'}\\
      \scm{H}^j_{Y_2} \circ \gamma &\neq \scm{H}^{j'}_{Y_2} \circ \gamma,
  \end{align}
  where, given the intervention map,
  the concrete interventions
  correspond
  to the following
  abstract interventions
  \begin{align}
  j &=
  ( \set{Y} \setminus \{Y_1, Y_2\} \gets \tau(\vec{v}), Y_1 \gets \tau_{Y_1}(\vec{c}))\\
  j' &=
  ( \set{Y} \setminus \{Y_1, Y_2\} \gets \tau(\vec{v}), Y_1 \gets \tau_{Y_1}(\vec{c}')).
  \end{align}
  Therefore,
  due to
  the surjectivity of $\gamma$,
  it also holds
  \begin{align}
      \scm{H}^j_{Y_2} \neq \scm{H}^{j'}_{Y_2}.
  \end{align}
  Consequently,
  since $j$ and $j'$
  differ only in $Y_1$
  and fix everything but $Y_2$,
  $Y_1$ has a direct effect on $Y_2$,
  i.e., $Y_1 \to Y_2$.
\end{proof}

\subsection{Corollary~\ref{lemma:corollarysufcon}}\label{proof:corollarysufcon}
\paragraph{\Cref{lemma:corollarysufcon} (Sufficient Directed Paths)}
  Let $\scm{H}$ be a $\mat{T}$-abstraction of $\scm{L}$,
  where $\scm{H}$ and $\scm{L}$
  are two linear SCMs
  respectively on variables $\set{Y}$ and $\set{X}$ with graphs $\mathcal{G}_\scm{H}$ and $\mathcal{G}_\scm{L}$.
  Then, 
  for any pair
  of relevant variables $X_1,X_2\in\Pi_R(\set{Y})$,
  such that
  $X_1\in\Pi_R(Y_1)$
  and
  $X_2\in\Pi_R(Y_2)$
  with $Y_1 \neq Y_2 \in \set{Y}$,
  it holds that
  \begin{align}
    X_1 \anc X_2 \ \ \mathrm{in} \ \mathcal{G}_\scm{L}
    \implies
    Y_1 \anc Y_2 \ \ \mathrm{in} \ \mathcal{G}_\scm{H}.
  \end{align}
\begin{proof}
   Given \Cref{lemma:sufcon},
   whenever there exists a $\mat{T}$-direct
   path between relevant variables $X_1\in\rel{Y_1}$
   and $X_2\in\rel{Y_2}$
   there must exist an abstract
   edge $Y_1 \to Y_2$.
   However, if the path is not $\mat{T}$-direct,
   then there must exists
   some relevant variable $X_3\in\rel{Y_3}$
   for another abstract variable $Y_3$ along the path.
   We firstly
   consider the case where $Y_3 \neq Y_1$ and $Y_3 \neq Y_2$.
   Consequently,
   there must exist an edge $Y_1 \to Y_3$ and,
   by applying the same argument on the path $X_3 \anc X_2$,
   the corollary holds for $Y_1 \anc Y_2$.
   Due to the acyclicity of the abstract graph,
   the case where $Y_3 = Y_1$ or $Y_3 = Y_2$
   can arise only at the beginning (resp. the end) of the path.
   In this case,
   we could consider the successive variable
   until we get one different from $Y_1, Y_2$, if any.
   If there is none, then there exists a $\mat{T}$-direct path between the relevant variables of $Y_1, Y_2$ and we fallback to the scenario of \Cref{lemma:sufcon},
   which directly entails $Y_1 \anc Y_2$.
\end{proof}

\subsection{Example~\ref{ex:faithfulness}}\label{proof:faithfulness}
\paragraph{\Cref{ex:faithfulness} (Unfaithful Concrete Model)}
\begin{proof}
To prove $\mat{T}$-abstraction
of the example, we anticipate
the parametrical characterization
of linear abstraction
which we introduce in \Cref{subsec:weightconc}.
In particular,
given the adjacencies
of the model,
\begin{align}
    \mat{W} &= \begin{bmatrix}
      0 & 1 & -1 & 0 & 1\\
      0 & 0 & 0 & 1 & 0\\
      0 & 0 & 0 & 1 & 0\\
      0 & 0 & 0 & 0 & 1\\
      0 & 0 & 0 & 0 & 0
    \end{bmatrix}\\
    \mat{M} &= \begin{bmatrix}
      0 & 0 & 1\\
      0 & 0 & 1\\
      0 & 0 & 0
    \end{bmatrix}
\end{align}
the necessary form for
the exogenous abstraction function,
which we will introduce in \Cref{lemma:exoabs},
is
\begin{align}
    \mat{S} = \begin{bmatrix}
      1 & 0& 0\\
      0 & 1& 0\\
      0 & 1& 0\\
      0 & 1& 0\\
      0 & 0& 1
    \end{bmatrix}.
\end{align}
Consequently,
we can prove abstraction by showing that
for any $Y_i, Y_j$
it holds that
\begin{align}
    \mat{W}_{ij}\vec{s}_j = m_{ij} \vec{t}_i.
\end{align}
For this example,
of particular interest is the case $Y_1 \to Y_2$,
where it holds that
\begin{align}
    \mat{W}_{1,2}\vec{s}_2 &= m_{1,2} \vec{t}_1\\
    \begin{bmatrix}
       1 & -1 & 0 
    \end{bmatrix}    
    \begin{bmatrix}
       1\\
       1\\
       1\\
    \end{bmatrix}    
    &=
    0 \cdot \begin{bmatrix}1\end{bmatrix}\\
    0 &= 0,
\end{align}
and thus $\scm{H}$ $\mat{T}$-abstracts $\scm{L}$.
\end{proof}

\subsection{Theorem~\ref{theo:connectivity}}\label{proof:connectivity}
\paragraph{\Cref{theo:connectivity} (Abstract Connectivity)}
  Let $\scm{H}$ be a $\mat{T}$-abstraction of $\scm{L}$,
  where $\scm{H}$ and $\scm{L}$
  are two linear SCMs
  respectively on variables $\set{Y}$ and $\set{X}$ with graphs $\mathcal{G}_\scm{H}$ and $\mathcal{G}_\scm{L}$.
  Then,
  there exists an edge ${Y_1 \to Y_2}$ in $\mathcal{G}_\scm{H}$
  if and only if
  for each $X_1\in\Pi_R(Y_1)$
  there exists $X_2\in\Pi_R(Y_2)$
  such that $X_1 \taudirect X_2$ in $\mathcal{G}_\scm{L}$.
\begin{proof}
  The sufficient condition follows immediately from \Cref{lemma:sufcon}, where we already proved that any $\mat{T}$-direct path between relevant variables entails an abstract edge.
  
  To prove the necessary condition,
  we consider instead two abstract interventions $j,j'$
  which differ only in $Y_1$
  and fix everything but $Y_2$.
  Formally,
  \begin{align}
     j  &= (Y_1 \gets k, \set{V} \gets \set{v})\\
     j' &= (Y_1 \gets k', \set{V} \gets \set{v}),
  \end{align}
  where $\set{V} = \set{Y} \setminus \{Y_1, Y_2\}$.
  Consequently,
  since $Y_1$ has a direct linear effect on $Y_2$,
  it holds that 
  \begin{align}\label{eq:proofnecessary}
      \scm{H}^j_{Y_2} &\neq \scm{H}^{j'}_{Y_2}\\
      \scm{H}^j_{Y_2} \circ \gamma &\neq \scm{H}^{j'}_{Y_2} \circ \gamma\\
     \tau_{Y_2} \circ \scm{L}^i &\neq \tau_{Y_2} \circ \scm{L}^{i'},
  \end{align}
  for any intervention $i, i'$ such that $\omega(i)=j$ and $\omega(i')=j'$.

  Let now $X_1 \in \Pi_R(Y_1)$
  be a relevant concrete variable
  for $Y_1$, and $t_{11}$ be the non-zero coefficient
  from $X_1$ to $Y_1$
  in the linear abstraction transformation~$\mat{T}$.
  We can then build two concrete interventions $i,i'$
  by setting all relevant variables of $Y_1$
  to zero, except for $X_1$.
  Formally,
  the interventions have the following form
  \begin{align}
      i &= (X_1 \gets \frac{k}{t_{11}},\, \Pi(Y_1) \setminus \{X_1\} \gets \vec{0},\, \ldots)\\
      i' &= (X_1 \gets \frac{k'}{t_{11}},\, \Pi(Y_1) \setminus \{X_1\} \gets \vec{0},\, \ldots).
  \end{align}
  If we suppose that
  it does not exist a variable $X_2\in\Pi_R(Y_2)$
  such that $X_1 \taudirect X_2$,
  all directed paths $X_1 \anc X_2$, if any,
  are mediated by a relevant variable
  of any abstract variable $Y\in \set{Y}\setminus \{Y_2\}$.
  Consequently, given our construction of $j,j'$
  and consequently $i,i'$,
  any path is mediated by an intervened variable
  and, therefore, it holds
  \begin{align}
     \tau_{Y_2} \circ \scm{L}^i = \tau_{Y_2} \circ \scm{L}^{i'},
  \end{align}
  which however breaks interventional consistency
  and implies that $\scm{H}$ is not
  a $\mat{T}$-abstraction of $\scm{L}$,
  proving the necessary condition by contradiction.
\end{proof}

\subsection{Corollary~\ref{cor:necessary}}\label{proof:necessary}
\paragraph{\Cref{cor:necessary} (Connectivity Violation)}
  Let $\scm{H}$ and $\scm{L}$
  be two linear SCMs
  respectively on variables $\set{Y}$ and $\set{X}$ with graphs $\mathcal{G}_\scm{H}$ and $\mathcal{G}_\scm{L}$. Consider a linear transformation~$\mat{T}$ between them leading to the sets of relevant variables $\Pi_R(\set{Y})$.
  If there exists three variables
  $X_1\in\Pi_R(Y_1)$,
  $X_2\in\Pi_R(Y_2)$, and
  $X_3\in\Pi_R(Y_1)$,
  such that
  both conditions hold
\begin{itemize}
    \item%
      $X_1 \taudirect X_2$
      in $\mathcal{G}_\scm{L}$, and
    \item%
      for any $X_4 \in \Pi_R(Y_2)$,
      $X_3 \centernot\taudirect X_4$
      is not in $\mathcal{G}_\scm{L}$,
\end{itemize}
    then $\scm{H}$ is not a $\mat{T}$-abstraction of $\scm{L}$.
\begin{proof}
Follows directly from \Cref{lemma:sufcon} which applied to the first item implies that $Y_1 \to Y_2$, and from \Cref{theo:connectivity}, which applied to the second item implies that $Y_1 \not \to Y_2$, hence providing a contradiction to the assumption that $\scm{H}$ is a $\mat{T}$-abstraction of $\scm{L}$.
\end{proof}

\subsection{Corollary~\ref{cor:exogabs}}\label{proof:exogabs}
\paragraph{\Cref{cor:exogabs} (Exogenous Abstraction)}
  Let $\scm{H}=(\set{Y},\set{U},\set{g},\dist{P}_{\set{U}})$ be a {$\mat{T}$-abstraction} of 
  $\scm{L}=(\set{X},\set{E},\set{f},\dist{P}_{\set{E}})$,
  where $\scm{H}$ and $\scm{L}$
  are two linear SCMs.
  Then, the exogenous abstraction function
  $\gamma\colon\dom{\set{E}}\to\dom{\set{U}}$,
  has form
  \begin{align}
      \gamma(\vec{e}) = \mat{S}^\tr \vec{e},
  \end{align}
  where $\mat{S} = \mat{F} \mat{T}\mat{G}^\inv$
  and $\mat{F}, \mat{G}$
  are
  the linear
  transformations
  of
  respectively
  the reduced forms of $\scm{L}$ and $\scm{H}$, i.e., $\scm{L}(\vec{e}) = \mat{F}^T \vec{e}$ and $\scm{H}(\vec{u}) = \mat{G}^T \vec{u}$.
\begin{proof}
    Since $\scm{H}$ $\mat{T}$-abstracts $\scm{L}$,
    it must hold $\tau\circ\scm{L}=\scm{H}\circ\gamma$.
    Consequently, due to the invertibility
    of the reduced form $\scm{H}$
    of linear SCMs, it holds that
    \begin{align}
        \gamma = \scm{H}^\inv\circ\tau\circ\scm{L}.
    \end{align}
    Since, $\scm{L}$, $\tau$, and $\scm{H}^\inv$
    are linear transformations,
    their composition coincides
    with a linear transformation
    $\mat{S}=\mat{F}\mat{T}\mat{G}^\inv$.
\end{proof}

\subsection{Lemma~\ref{lemma:blockordering}}\label{proof:blockordering}
\paragraph{\Cref{lemma:blockordering} (Block Composition)}
  Let $\scm{H}$ be a $\mat{T}$-abstraction of $\scm{L}$,
  where $\scm{H}$ and $\scm{L}$
  are two linear SCMs
  respectively on variables $\set{Y}$ and $\set{X}$.
  Then,
  for any abstract variable $Y\in\set{Y}$,
  it holds
  $X\in\Pi(Y)$ if and only if
  \begin{itemize}
      \item $X\in\Pi_R(Y)$, or
      \item  $X \not \in \Pi_R(\set{Y})$, i.e., $X$ is irrelevant, and there exists $X'\in\Pi_R(Y)$ s.t. $X \taudirect X'$.
  \end{itemize}
\begin{proof}
  Let $Y$ be an abstract variable
  and ${j = (\pa{Y} \gets \vec{k})}$
  be a hard intervention
  fixing all of its endogenous parents.
  Consequently,
  the value of the abstract variable,
  $\scm{H}^j_Y(\vec{u})$ depends only
  on its exogenous term $U_Y$.
  Further,
  given the definition of concrete block,
  the formulation
  \begin{align}
  \scm{H}^j_Y(\gamma(\vec{e})) &=
  \scm{H}^j_Y(\mat{S}^\tr \vec{e}),
  \end{align}
  depends only on the exogenous
  terms $\vec{e}_{\Pi(Y)}$.
  Therefore, given the interventional consistency property
  \begin{align}
  \scm{H}^j_Y(\gamma(\vec{e})) &= \tau_Y(\scm{L}^i_{\Pi_R(Y)}(\vec{e})),
  \end{align}
  and the lack of cancelling paths,
  $\scm{L}^i_{\Pi_R(Y)}$
  also depends only on the exogenous terms $\vec{e}_{\Pi(Y)}$,
  for any concrete intervention
  \begin{align}
    i = (\Pi_R(\pa{Y}) \gets \vec{c}),
  \end{align}
  where $\tau_{\pa{Y}}(\vec{c}) = \vec{k}$.
  Notably,
  given the intervention $i$,
  the structural mechanisms of $\Pi_R(Y)$
  depend only on the exogenous noise
  of the relevant variables
  and on those variables
  whose direct path
  is non-mediated by another relevant variable.
  Given \Cref{lemma:sufcon},
  any of such relevant variables
  must be in the relevant set of a parent,
  and thus be constrained by the intervention $i$.
  Consequently,
  $\scm{L}^i_{\Pi_R(Y)}$
  depends only on its relevant variables
  and the \emph{irrelevant} variables
  with a $\mat{T}$-direct path
  towards the former.
\end{proof}

\subsection{Lemma~\ref{lemma:disjointconstitutive}}\label{proof:disjointconstitutive}
\paragraph{\Cref{lemma:disjointconstitutive} (Disjoint Block)}
  Let $\scm{H}$ be a $\mat{T}$-abstraction of $\scm{L}$,
  where $\scm{H}$ and $\scm{L}$
  are two linear SCMs
  respectively on variables $\set{Y}$ and $\set{X}$.
  If for any two distinct endogenous variables $Y_1, Y_2$
  it holds that $\Pi(Y_1) \cap \Pi(Y_2) \neq \emptyset$, then 
  the abstract model is not causally sufficient.
\begin{proof}
  By definition of concrete block~(\Cref{def:block}),
  each abstract exogenous term $U_Y$
  is a function $\gamma$
  of the noise terms of the block $\Pi(Y)$.
  Therefore,
  given two variables $Y_1, Y_2\in\set{Y}$,
  we can write
  \begin{align}
  U_1 &= \gamma_{1} ( E_{\Pi(Y_1)})\\
  U_2 &= \gamma_{2} ( E_{\Pi(Y_2)}).
  \end{align}
  Therefore, 
  whenever the blocks share
  a subset of variables ${\set{S}} = \Pi(Y_1) \cap \Pi(Y_2)$,
  both $U_1$ and $U_2$
  are a function of the exogenous terms
  \begin{align}
    \set{V} =
    \{ E_X \in \set{E}
    \mid X \in \set{S}\}.
  \end{align}
  Consequently,
  the exogenous terms $U_1, U_2$
  are not independent and
  the variables $Y_1, Y_2$
  are then confounded.
\end{proof}

\subsection{Theorem~\ref{theorem:absord}}\label{proof:absord}
\paragraph{\Cref{theorem:absord} (Block Ordering)}
  Let $\scm{H}$ be a $\mat{T}$-abstraction of $\scm{L}$,
  where $\scm{H}$ and $\scm{L}$
  are two linear SCMs
  respectively on variables $\set{Y}$ and $\set{X}$ with graphs $\mathcal{G}_\scm{H}$ and $\mathcal{G}_\scm{L}$.
  Then,
  for any valid topological ordering $\prec_{\scm{H}}$ of $\mathcal{G}_\scm{H}$
  there exists a valid ordering $\prec_{\scm{L}}$ of $\mathcal{G}_\scm{L}$
  such that
  for any $Y_1, Y_2, Y \in \set{Y}$:
  \begin{itemize}
      \item%
      $Y_1 \prec_{\scm{H}} Y_2 \iff
      \Pi(Y_1) \prec_{\scm{L}} \Pi(Y_2)$, and
      \item%
      $\Pi(Y) \prec_{\scm{L}} \big( \set{X} \setminus \Pi (\set{Y}) \big). %
        $
  \end{itemize}
\begin{proof}

Firstly, we recall
that in a valid topological order,
a variable precedes another
only if
there is a directed path
from the former to the latter~\citep{bondy_graph_2008}.
\begin{align}
    X_1 \anc X_2 &\implies X_1 \prec X_2
\end{align}
Since we always compare abstract variables
with abstract variables
and concrete variables with concrete variables,
in the following
we ease the notation
by avoiding the subscript on the precedence operator~$\prec$.

We show the existence
of a valid topological ordering
on the concrete model by construction.
Given the topological ordering on the abstract model,
we assign to each abstract node~$Y\in\set{Y}$
an integer $\rho_{\set{Y}}(Y) \in \{1, \ldots, |\set{Y}|\}$
such that
\begin{align}
    Y_1 \prec Y_2 \iff \rho_{\set{Y}}(Y_1) < \rho_{\set{Y}}(Y_2).
\end{align}
Then, we can take any valid topological
ordering within any concrete block $\Pi(Y)$
and assign in the same way $\rho_{\Pi(Y)}(X)$
for any
$Y\in\set{Y}$
and
$X\in\Pi(Y)$.
We do the same
for the set~$\set{Q}$
of concrete variables
outside of any block,
which we formally define as follows
\begin{align}
    \set{Q} = \set{X} \setminus \bigcup_{Y\in\set{Y}} \Pi(Y).
\end{align}
We then assign
the ``position''
of each
concrete variable $X\in\set{X}$
through a further integer
defined as follows,
\begin{align}
    \rho_{\set{X}} = \begin{cases}
        \sum_{Y' \prec Y} |\Pi(Y')| + \rho_{\Pi(Y)}(X)
        & \exists Y .\, X \in \Pi(X)\\
        \sum_{Y \in\set{Y}} |\Pi(Y)| + \rho_{\set{Q}}(X)
        & X \in \set{Q}.
    \end{cases}
\end{align}
Notably, since the blocks do not overlap (\Cref{lemma:disjointconstitutive}),
the assignment is unique.
We finally define the concrete topological ordering
for any $X_1, X_2 \in\set{X}$ as
\begin{align}
  X_1 \prec X_2 \iff \rho_{\set{X}}(X_1) < \rho_{\set{X}}(X_2).  
\end{align}

Given this ordering, it holds by construction that
\begin{align}
  \forall Y_1, Y_2 \in \set{Y}.\, 
  Y_1 \prec_{\scm{H}} Y_2 \iff
  \Pi(Y_1) \prec_{\scm{L}} \Pi(Y_2)\\
  \forall Y \in \set{Y}.\,
    \Pi(Y) \prec_{\scm{L}} \{X\in\set{X} \mid X \notin \bigcup_{Y\in\set{Y}} \Pi(Y)\}.
\end{align}
Therefore,
to finally prove the Theorem we have to show
that the ordering we defined is valid for the concrete graph.
Formally,
we have to show that, for any $X_1, X_2\in\set{X}$,
\begin{align}
    X_1 \to X_2 &\implies X_1 \prec X_2\\
                &\implies \rho_{\set{X}}(X_1) < \rho_{\set{X}}(X_2).
\end{align}
\textit{Case $\{X_1, X_2\} \subset\Pi(Y) \lor \{X_1, X_2\}\subset\set{Q}$.}
Whenever $X_1 \to X_2$ and $X_1, X_2$ are in the same block $\Pi(Y)$
for some $Y\in\set{Y}$ or are both in $\set{Q}$,
then $\rho_{\set{X}}(X_1) < \rho_{\set{X}}(X_2)$
by definition.\\
\textit{Case $X_1\in\Pi(Y_1), X_2\in\set{Q}$.}
Also holds by definition.\\
\textit{Case $X_1\in\set{Q}, X_2\in\Pi(Y)$.}
By definition of block,
this case never occurs,
since otherwise $X_1$ would be in $\Pi$ (\Cref{lemma:blockordering}).
\textit{Case $X_1\in\Pi(Y_1), X_2\in\Pi(Y_2)$.}
Further,
whenever $X_1 \to X_2$ such that $X_1\in\Pi(Y_1)$
for some $Y_1$ and $X_2\in\Pi(Y_2)$ for some $Y_2$,
then $X_1$ is relevant, otherwise it would have also been
in the block $\Pi(Y_2)$, which are necessarily disjoint (\Cref{lemma:disjointconstitutive}).
Therefore, given the sufficient condition
on the existence of an abstract edge (\Cref{lemma:sufcon}),
it must hold
\begin{align}
Y_1 &\to Y_2 \\
\implies Y_1 &\prec Y_2 \\
\implies \Pi(Y_1) &\prec \Pi(Y_2) \\
\implies X_1 &\prec X_2.
\end{align}
\end{proof}

\subsection{Lemma~\ref{lemma:ignvar}}\label{proof:ignvar}
\paragraph{\Cref{lemma:ignvar} (Submodel Abstraction)}
  Let $\scm{H}$ and $\scm{L}$
  be two linear SCMs
  respectively on variables $\set{Y}$ and $\set{X}$.
  Then,
  $\scm{H}$ is a $\mat{T}$-abstraction of $\scm{L}$
  if and only if
  $\scm{H}$ is a $\mat{T}$-abstraction of $\scm{L}^\prime$,
  where $\scm{L}^\prime$ is a submodel of $\scm{L}$
  defined on the subset of variables
  $\set{X}^\prime = \Pi(\set{Y})$, i.e., all of the variables in the concrete blocks.
\begin{proof}
    The Lemma directly follows from \Cref{theorem:absord},
    where the variables not in any block
    always follow in the topological ordering
    the remaining. Therefore, by removing them,
    for any intervention $i$
    the interventional consistency
    $\tau \circ \scm{L}'^i  = \tau \circ \scm{L}^i$
    still holds
    since they do not influence any relevant variable, hence the abstraction function~$\tau$, nor any block, hence the exogenous abstraction function~$\gamma$.
    Similarly, we could add as many variables
    and mechanism not influencing the blocks
    and interventional consistency would still hold. 
\end{proof}

\subsection{Lemma~\ref{lemma:exoabs}}\label{proof:exoabs}
\paragraph{\Cref{lemma:exoabs} (Exogenous Abstraction)}
  Let $\scm{H}=(\set{Y}, \set{U}, \mat{M},  \dist{P}_{\set{U}})$
  and $\scm{L}=(\set{X},  \set{E}, \mat{W}, \dist{P}_{\set{E}})$
  be two linear SCMs
  such that $\scm{H}$
  is a $\mat{T}$-abstraction
  of $\scm{L}$, such that $\mat{W}$ follows permutation $\pi_{\scm{H}}$.
  Then, the exogenous abstraction function $\gamma\colon\dom{\set{E}}\to\dom{\set{U}}$
  is unique and
  has form ${\gamma(\vec{e}) = \mat{S}^\tr \vec{e}}$
  for a linear transformation ${\mat{S}\in\real^{d \times b}}$
  defined as the upper-diagonal block matrix
  \begin{align}
    \mat{S} = \begin{bmatrix}
      \vec{s}_1 & \mat{0} & \cdots & \mat{0} \\
      \mat{0} & \vec{s}_2 & \cdots & \mat{0} \\
      \vdots & \vdots & \ddots & \vdots \\
      \mat{0} & \mat{0} & \cdots & \vec{s}_b,
    \end{bmatrix}
  \end{align}
  where
  $\vec{s}_k = \mat{F}_{kk} \vec{t}_k = {(\mat{I} - \mat{W}_{kk})}^\inv\vec{t}_k$
  for any $Y_k\in\set{Y}$.%
\begin{proof}
  Given the definition of $\mat{T}$-abstraction,
  we can rephrase observational consistency as
  \begin{align}
    \tau \circ \scm{L} &= \scm{H} \circ \gamma\\
    \mat{FT} &= \mat{SG}
  \end{align}
  where $\mat{F}$ and $\mat{G}$
  are respectively the reduced
  forms of the concrete and the abstract SCM\@.
  Consequently,
  by exploiting the block-definition of $\mat{T}$,
  we can reformulate the left side of the equation as
  \begin{align}
    \begin{bmatrix}
      \mat{F}_{11} & \mat{F}_{12} & \cdots & \mat{F}_{1b} \\
      \mat{0} & \mat{F}_{22} & \cdots & \mat{F}_{2b} \\
      \vdots & \vdots & \ddots & \vdots \\
      \mat{0} & \mat{0} & \cdots & \mat{F}_{bb}
    \end{bmatrix}
    \begin{bmatrix}
      \vec{t}_1 & \vec{0} & \cdots & \vec{0} \\
      \vec{0} & \vec{t}_2 & \cdots & \vec{0} \\
      \vdots & \vdots & \ddots & \vdots \\
      \vec{0} & \vec{0} & \cdots & \vec{t}_b
    \end{bmatrix}\\
    =\begin{bmatrix}
      \mat{F}_{11}\vec{t}_1 & \mat{F}_{12}\vec{t}_2 & \cdots & \mat{F}_{1b}\vec{t}_b \\
      \mat{0} & \mat{F}_{22}\vec{t}_2 & \cdots & \mat{F}_{2b}\vec{t}_b \\
      \vdots & \vdots & \ddots & \vdots \\
      \mat{0} & \mat{0} & \cdots & \mat{F}_{bb}\vec{t}_b
    \end{bmatrix}
  \end{align}
  Given that
  block variables are not shared (\Cref{lemma:disjointconstitutive})
  and follow the same topological order of $\mat{T}$,
  the exogenous transformation must also have form
  \begin{align}
    \mat{S} = \begin{bmatrix}
      \vec{s}_1 & \mat{0} & \cdots & \mat{0} \\
      \mat{0} & \vec{s}_2 & \cdots & \mat{0} \\
      \vdots & \vdots & \ddots & \vdots \\
      \mat{0} & \mat{0} & \cdots & \vec{s}_b,
    \end{bmatrix}.
  \end{align}
  We can therefore reformulate the right
  side $\mat{SG}$
  of the observational consistency
  equation as
  \begin{align}
      \begin{bmatrix}
      \vec{s}_1 & \mat{0} & \cdots & \mat{0} \\
      \mat{0} & \vec{s}_2 & \cdots & \mat{0} \\
      \vdots & \vdots & \ddots & \vdots \\
      \mat{0} & \mat{0} & \cdots & \vec{s}_b,
    \end{bmatrix}
    \begin{bmatrix}
      1 & g_{12} & \cdots & g_{1b} \\
      0 & 1 & \cdots & g_{2b} \\
      \vdots & \vdots & \ddots & \vdots \\
      0 & 0 & \cdots & 1
    \end{bmatrix}\\
    =\begin{bmatrix}
      \vec{s}_1 & g_{12}\vec{s}_1 & \cdots & g_{1b}\vec{s}_1 \\
      \mat{0} & \vec{s}_2 & \cdots & g_{2b}\vec{s}_2 \\
      \vdots & \vdots & \ddots & \vdots \\
      \mat{0} & \mat{0} & \cdots & \vec{s}_b.
    \end{bmatrix}
  \end{align}
  Consequently, for any $Y_i\in\set{Y}$,
  it holds $\vec{s}_i = \mat{F}_{ii}\vec{t}_i$.
\end{proof}

\subsection{Theorem~\ref{theo:concretization}}\label{proof:concretization}
\paragraph{\Cref{theo:concretization} (Block Abstraction)}
  Let $\scm{H}=(\set{Y}, \set{U}, \mat{M}, \dist{P}_{\set{U}})$
  and $\scm{L}=(\set{X}, \set{E}, \mat{W}, \dist{P}_{\set{E}})$
  be two linear SCMs with graphs $\mathcal{G}_{\scm{H}}$ and $\mathcal{G}_{\scm{L}}$ respectively.
  Then $\scm{H}$ is a linear $\mat{T}$-abstraction of $\scm{L}$
  if and only if
  for any valid topological ordering $\prec_{\scm{H}}$ of $\mathcal{G}_{\scm{H}}$
  there exists a valid ordering $\prec_{\scm{L}}$ of $\mathcal{G}_{\scm{L}}$
  such that,
  for any $Y_i,Y_j\in\set{Y}$ it holds
  \begin{align}\label{eq:weightconsistency}
    Y_i \prec_{\scm{H}} Y_j &\iff \Pi(Y_i) \prec_{\scm{L}} \Pi(Y_j), \ \mathrm{and}\\
    \mat{W}_{ij}\vec{s}_j &= m_{ij} \vec{t}_i,
  \end{align}
  where $\mat{W}_{ij}$ is the $i$-th element on the $j$-th column of $\mat{W}$, and $m_{ij}$ is the $i$-th element on the $j$-th column of $\mat{M}$.
\begin{proof}
Firstly, we introduce
the following decomposition
of the reduced forms
of the concrete and the abstract model,
which we separately prove in \Cref{proof:decomposition}.
\begin{align}
\mat{F}_{ij} &= \begin{cases}
  {(\mat{I} - \mat{W}_{ii})}^\inv & \text{if } i=j\\
  \mat{F}_{ii} (\mat{W}_{ij} + \mat{R}_{ij}) \mat{F}_{jj} & \text{if } i<j\\
  \mat{0} & \text{otherwise},\\
\end{cases}\\
\mat{R}_{ij} &= \sum_{i < k < j} \mat{W}_{ik} \mat{F}_{kk} (\mat{W}_{kj} + \mat{R}_{kj})\\
g_{ij} &= \begin{cases}
  1 & \text{if } i=j\\
  m_{ij} + \rho_{ij} & \text{if } i<j\\
  0 & \text{otherwise},\\
\end{cases}\\
\rho_{ij} &= \sum_{i < k < j} m_{ik} (m_{kj} + \rho_{kj})
\end{align}
\textit{Necessary Condition.}
We show that $\mat{T}$-abstraction
implies both conditions.
For the existence of a valid concrete ordering,
we invite the reader to consult the proof
of \Cref{theorem:absord}.
Therefore, we focus on proving that $\mat{T}$-abstraction
entails $\mat{W}_{ij} \vec{s}_j = m_{ij} \vec{t}_i$
for any $Y_i, Y_j \in \set{Y}$.
Given the decomposition
consistency condition $\mat{FT}=\mat{SG}$
from the proof of \Cref{lemma:exoabs},
for each $i<j$, it must hold that
\begin{align}
\mat{F}_{ij} \vec{t}_j
&= \vec{s}_{i} g_{ij}\\
\mat{F}_{ii} (\mat{W}_{ij} + \mat{R}_{ij}) \mat{F}_{jj} \vec{t}_j
&= \mat{F}_{ii} \vec{t}_i (m_{ij} + \rho_{ij})\\
(\mat{W}_{ij} + \mat{R}_{ij}) \vec{s}_j
&= \vec{t}_i (m_{ij} + \rho_{ij})\\
\mat{W}_{ij} \vec{s}_j
&= m_{ij} \vec{t}_i,
\end{align}
where the first step comes from
the previously introduced decomposition,
proved in \Cref{proof:decomposition}.
To prove the last step
we firstly notice that
\begin{align}
\mat{R}_{ij} \vec{s}_j = \rho_{ij}\vec{t}_i 
\iff
\mat{W}_{ij} \vec{s}_j = m_{ij} \vec{t}_i.
\end{align}
We then prove the statement for each row
by induction on the columns.
We take $j=i+1$ as base case,
where it holds
\begin{align}
    \mat{R}_{ij}\vec{s}_j &= \rho_{ij} \vec{t}_{i}\\
    \mat{0}\vec{s}_j &= 0 \cdot \vec{t}_{i}\\
    \vec{0} &= \vec{0}\\
    \implies
    \mat{W}_{ij} \vec{s}_j &= m_{ij} \vec{t}_i.
\end{align}
Consequently,
we can show that
\begin{align}
    &\mat{R}_{ij} \vec{s}_j =
    \sum_{i < k < j}
    \mat{W}_{ik} \mat{F}_{kk} (\mat{W}_{kj} + \mat{R}_{kj})
    \vec{s}_j\\
    &=\sum_{i < k < j}
    \mat{W}_{ik} \mat{F}_{kk} \mat{W}_{kj}
    \vec{s}_j +
    \mat{W}_{ik} \mat{F}_{kk} \mat{R}_{kj}
    \vec{s}_j\\
    &=\sum_{i < k < j}
    \mat{W}_{ik} \mat{F}_{kk} \mat{W}_{kj}
    \vec{s}_j +
    \mat{W}_{ik} \mat{F}_{kk} \rho_{kj}
    \vec{t}_k\\
    &=\sum_{i < k < j}
    \mat{W}_{ik} \mat{F}_{kk} m_{kj}
    \vec{t}_k +
    \mat{W}_{ik} \mat{F}_{kk} \rho_{kj}
    \vec{t}_k\\
    &=\sum_{i < k < j}
    \mat{W}_{ik} \mat{F}_{kk}
    \vec{t}_k
    m_{kj} +
    \mat{W}_{ik} \mat{F}_{kk}
    \vec{t}_k
    \rho_{kj}\\
    &=\sum_{i < k < j}
    m_{ik}\vec{t}_i m_{kj} +
    m_{ik}\vec{t}_i \rho_{kj}\\
    &=\sum_{i < k < j}
    m_{ik} (m_{kj} + \rho_{kj}) \vec{t}_i\\
    &= \rho_{ij} \vec{t}_i.
\end{align}
\textit{Sufficient Condition.}
We now show that the conditions imply
interventional consistency of the abstraction.
That is, we want to prove that
\begin{align}
    \tau_Y \circ \scm{L}^\iota_{\Pi(Y)}
    &= \scm{H}^{\omega(\iota)}_Y \circ \gamma,
\end{align}
for any concrete intervention $\iota$
on the relevant sets
defined by the linear abstraction transformation $\mat{T}$.
Firstly, we notice that the equation
is immediately true for any abstract variable $Y\in\set{Y}$
whenever the intervention targets its relevant set.
Therefore,
we focus on the case
where the abstract intervention $\omega(\iota)$
does not affect $Y$.
Consequently,
given that we assume that the
topological ordering of the blocks
coincides with that of the abstract variables,
we can decompose the concrete model as
\begin{align}
    \scm{L}^\iota_{\Pi(Y_j)}(\vec{e})
    &=
    \sum_{Y_i\in\pa{Y_j}}
    \left(
    {\left[
    \scm{L}^\iota_{\Pi(Y_i)}(\vec{e})
    \right]}^\tr
    \mat{W}_{ij} + \vec{e}^\tr_{\Pi(Y_j)}
    \right) \mat{F}_{jj},
\end{align}
where we
(i.) compute the linear contribution of the parents,
(ii.) sum the exogenos noise of the block,
(iii.) and apply the submodel composed of the internal connections in the block.
Similarly, we can decompose the abstract model as
\begin{align}
    \scm{H}_{Y_j}^{\omega(\iota)}(\vec{u})
    &=
    \sum_{Y_i\in\pa{Y_j}}
    \scm{H}_{Y_i}^{\omega(\iota)}(\vec{u})
    \cdot
    m_{ij}
    + u_j.
\end{align}
Abstraction holds whenever
interventional consistency is satisfied
by at least an exogenous transformation $\gamma$.
To continue the proof, we then define it as the linear transformation
from \Cref{lemma:exoabs}, where $\vec{s}_j = \mat{F}_{jj}\vec{t}_j$ for any $Y_j\in\set{Y}$.
Therefore,
we can reformulate interventional consistency as
\begin{align}
\begin{split}
    \sum_{Y_i\in\pa{Y_j}}
    \left(
    {\left[
    \scm{L}^\iota_{\Pi(Y_i)}(\vec{e})
    \right]}^\tr
    \mat{W}_{ij} + \vec{e}^\tr_{\Pi(Y_j)}
    \right) \mat{F}_{jj} \vec{t}_j
    \\= 
    \sum_{Y_i\in\pa{Y_j}}
    \scm{H}_{Y_i}^{\omega(\iota)}(\mat{S}^\tr \vec{e})
    \cdot
    m_{ij}
    + \vec{e}_{\Pi(Y_j)}^\tr \vec{s}_j,
\end{split}
\end{align}
which further simplifies to
\begin{align}
\begin{split}
    &\sum_{Y_i\in\pa{Y_j}}
    {\left[
    \scm{L}^\iota_{\Pi(Y_i)}(\vec{e})
    \right]}^\tr
    \mat{W}_{ij} \mat{F}_{jj} \vec{t}_j
    \\= 
    &\sum_{Y_i\in\pa{Y_j}}
    \scm{H}_{Y_i}^{\omega(\iota)}(\mat{S}^\tr \vec{e})
    \cdot
    m_{ij}
\end{split}
\end{align}
given our choice of the exogenous transformation $\mat{S}$.
We prove this last equation by induction
on the topological ordering
of the abstract graph.
In fact, as a base case,
for any root of the graph
the equation holds
given that the parent set is the empty set.
Consequently,
we can finally show
that 
$\mat{W}_{ij}\vec{s}_j = m_{ij} \vec{t}_i$
implies abstraction
as follows
\begin{align}
    &\sum_{Y_i\in\pa{Y_j}}
    \scm{H}_{Y_i}^{\omega(\iota)}(\mat{S}^\tr \vec{e})
    \cdot
    m_{ij}\\
    =
    &\sum_{Y_i\in\pa{Y_j}}
    {\left[
    \scm{L}^\iota_{\Pi(Y_i)}(\vec{e})
    \right]}^\tr \vec{t}_i
    \cdot
    m_{ij}\\
    =
    &\sum_{Y_i\in\pa{Y_j}}
    {\left[
    \scm{L}^\iota_{\Pi(Y_i)}(\vec{e})
    \right]}^\tr
    \mat{W}_{ij}
    \vec{s}_{j}\\
    =
    &\sum_{Y_i\in\pa{Y_j}}
    {\left[
    \scm{L}^\iota_{\Pi(Y_i)}(\vec{e})
    \right]}^\tr
    \mat{W}_{ij}
    \mat{F}_{jj}
    \vec{t}_{j}.
\end{align}
\end{proof}

\subsection{Model Reduction Decomposition}\label{proof:decomposition}
In the following,
we prove the decomposition
of the model reduction matrix~$\mat{F}$
from the proof in \Cref{proof:concretization}.
To simplifiy the notation, we define the matrix $\mat{A}=(\mat{I}-\mat{W})$.

\begin{proof}
Back-substituting
to solve $\mat{FA}=\mat{I}$
leads to
\begin{align}
  {\mat{F}}_{ij} = \begin{cases}
    \mat{A}_{ii}^\inv & i =j\\
    - \sum_{i<k\leq j} \mat{F}_{ii} \mat{A}_{ik} {\mat{F}}_{kj} & i <j\\
    0 & i > j
  \end{cases}.
\end{align}
Therefore, we want to prove that whenever $i<j$,
it holds
\begin{align}
  - \sum_{i<k\leq j} \mat{F}_{ii} \mat{A}_{ik} {\mat{F}}_{kj}
  =
  \mat{F}_{ii}(\mat{W}_{ij} + \mat{R}_{ij})\mat{F}_{jj},
\end{align}
where
\begin{align}
  \mat{R}_{ij} =
  \sum_{i < k < j} \mat{W}_{ik} \mat{F}_{kk} (\mat{W}_{kj} + \mat{R}_{kj}).
\end{align}

Overall,
we simplify the thesis
as follows
\begin{align}
  \mat{F}_{ii}(\mat{W}_{ij} + \mat{R}_{ij})\mat{F}_{jj}
  &=
  - \sum_{i<k\leq j} \mat{F}_{ii} \mat{A}_{ik} {\mat{F}}_{kj}
  \\
  \mat{F}_{ii}(\mat{W}_{ij} + \mat{R}_{ij})\mat{F}_{jj}
  &=
  \sum_{i<k\leq j} \mat{F}_{ii} \mat{W}_{ik} {\mat{F}}_{kj}
  \\
  \mat{F}_{ii}\mat{R}_{ij}\mat{F}_{jj}
  &=
  \sum_{i<k<j} \mat{F}_{ii} \mat{W}_{ik} {\mat{F}}_{kj}
  \\
  \mat{R}_{ij}\mat{F}_{jj}
  &=
  \sum_{i<k<j} \mat{W}_{ik} {\mat{F}}_{kj}.
\end{align}

We finally prove our thesis
by induction
on the decreasing row component~$i$,
starting
from $i=j-1$.
In the base case,
both sides of the equation
reduce to zero
and thus the statement holds.
We then prove the inductive case
by showing that if the statement
holds for any $k>i$, then it also
holds for $i$.
Formally,
\begin{align}
  &\sum_{i<k<j} \mat{W}_{ik} {\mat{F}}_{kj}\\
  &=
  \sum_{i<k<j} \mat{W}_{ik} \mat{F}_{kk} (\mat{W}_{kj} + \mat{R}_{kj}) \mat{F}_{jj}\\
  &=\mat{R}_{ij} \mat{F}_{jj}.
\end{align}
\end{proof}

\section{Dataset}\label{app:dataset}

\begin{figure}[H]
    \centering
    \includegraphics[width=\textwidth]{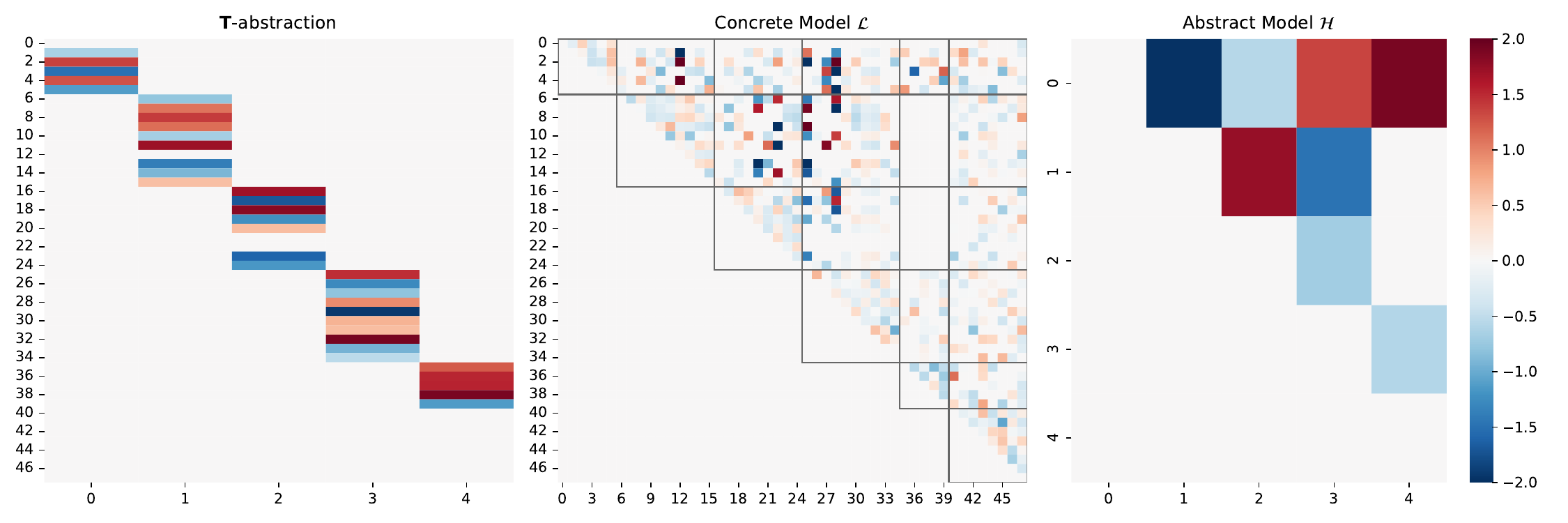}
    \caption{%
    Visualization of a pair of concrete-abstract models
    and their abstraction function.
    The abstract graph has 5 nodes and 8 edges
    while the concrete has 5 blocks of random size from $[5, 10]$,
    with an additional block for the ignored variables.
    }\label{fig:visualization}
\end{figure}

In the following,
we report further details
on the simulation procedure
used to generate the dataset
for the experiments,
which we also visualize in \Cref{fig:visualization}.

\paragraph{Abstract Model.}
Given a number of desired nodes
and edges,
we sample the abstract model
by randomly sampling
an Erd\H{o}s-R\'enyi graph
for the given parameters.
Then, we sample the weights
of the edges
from the uniform distribution
in the interval $[-2, -0.5] \cup [0.5, 2]$.

\paragraph{Abstraction Function.}
Given the abstract model,
we sample the abstraction function
by firstly assigning a block size to each node
from the uniform distribution,
whose minimum and maximum values
are given as input.
Then, within each block we randomly choose
at least half of the nodes to be \emph{relevant}
and randomly assign the remaining
as relevant or not.
We also sample a further block to contain the \emph{ignored} variables,
for which the abstraction function maps to zero.
We finally sample the abstraction coefficients
from the uniform distribution
in the interval $[-2, -0.5] \cup [0.5, 2]$.

\paragraph{Concrete Model.}
Given an abstract model and an abstraction function,
we sample the concrete model
using the algorithm in \Cref{alg:samplingblocks}.
Firtsly, we sample
the causal relations within each block
by randomly sampling
an upper triangular matrix
with non-zero entries
from the standard normal distribution.
Then,
we employ the Dirichlet distribution
to sample each vector $\vec{v}$
with sum one as requested by the algorithm
to explore the right-inverses
of the exogenous abstraction function.
Finally,
we randomly sample
from the standard normal distribution.
the weights
to connect ignored variables.

\paragraph{Data Generation.}
As we detailed in the main body,
we sample the data from the concrete model
by first sampling the non-Gaussian noise
and then by abstracting the noise
to sample from the abstract model.
In all experiments, we use the Exponential distribution.
We then normalize the data to have zero mean and unit variance
and permute all the variables
in both the concrete and abstract samples.

\section{Additive Noise on Abstract Observations}\label{app:noisy}

\begin{figure}
  \centering
  \begin{subfigure}{\linewidth}
    \centering
    \includegraphics[width=\linewidth]{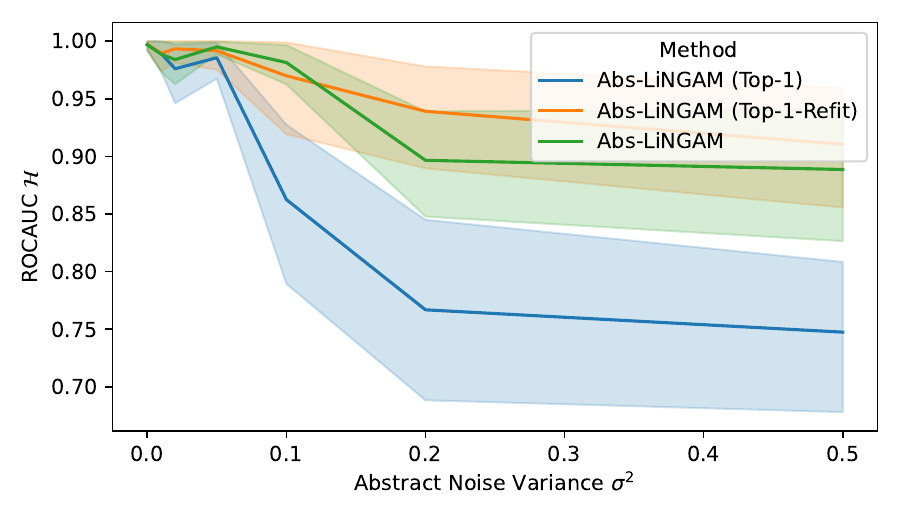}
  \end{subfigure}
  \begin{subfigure}{\linewidth}
    \includegraphics[width=\linewidth]{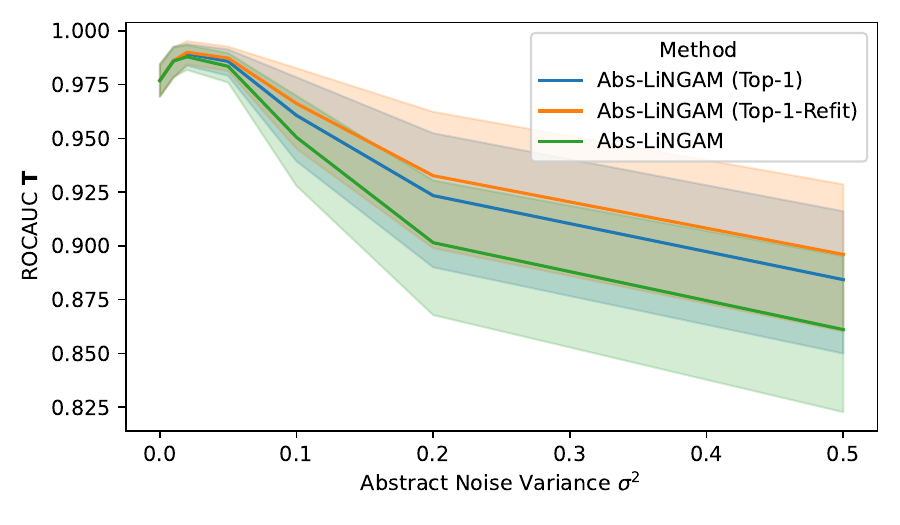}
  \end{subfigure}
  \begin{subfigure}{\linewidth}
    \includegraphics[width=\linewidth]{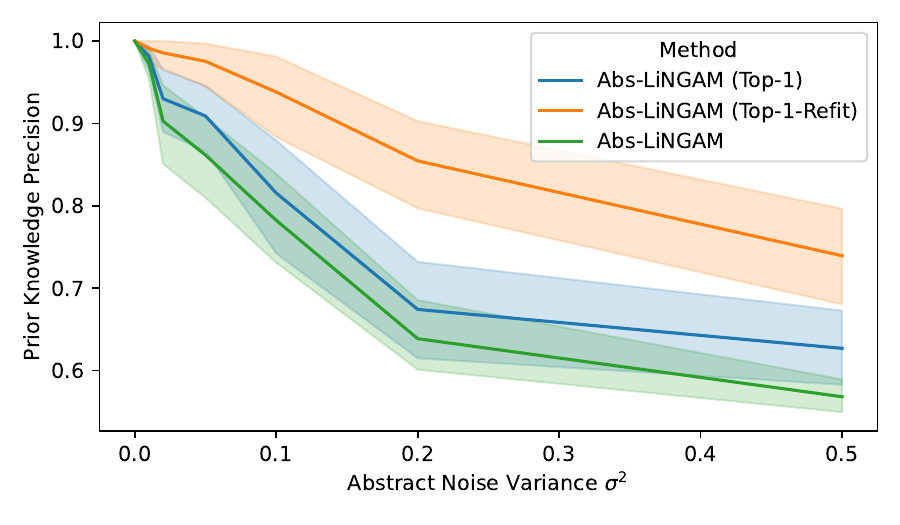}
  \end{subfigure}
  \begin{subfigure}{\linewidth}
    \includegraphics[width=\linewidth]{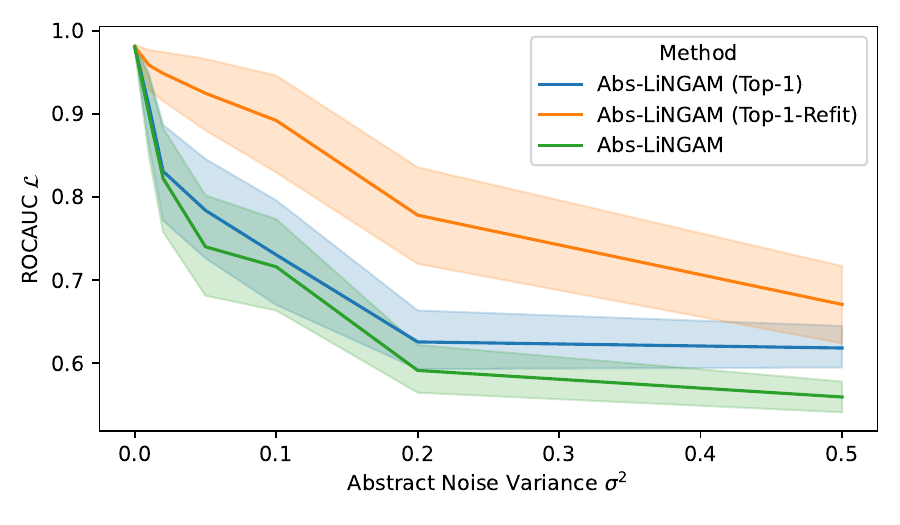}
  \end{subfigure}
  \caption{%
    Results of Abs-LiNGAM
    over pairs of abstract ($b=5$ nodes)
    and concrete ($d\in[25,50]$ nodes)
    linear SCMs
    after perturbing the abstract observations
    with normal noise
    of increasing variance $\sigma^2$.
    We denote as ``Top-1'' the strategy
    where we force the selection
    of at most a single abstract variable
    per concrete one and as ``Top-1-Refit''
    the one where we then refit each abstraction vector.
    All results are averaged over 30 independent runs
    with $|\dset_{\scm{L}}| = 20000$ concrete samples
    and $|\dset_{\scm{J}}| = 150$ paired samples.
  }\label{fig:noisy}
\end{figure}

In this section, we discuss strategies to handle a further scenario where we consider abstract observations to be further perturbed by random noise. We consider the following generative model for the abstract observations:
\begin{align}
  \vec{e}^{(i)} &\sim \operatorname{Exponential}
                &\text{for } i=1,\dots,|\dset_{\scm{L}}|,\\
  \vec{x}^{(i)} &= \scm{L}(\vec{e}^{(i)})
                &\text{for } i=1,\dots,|\dset_{\scm{L}}|,\\
  \vec{y}^{(i)} &= \scm{H}(\gamma(\vec{e}^{(i)})) + \vec{\epsilon}^{(i)}
                &\text{for } i=1,\dots,|\dset_{J}|,
\end{align}
where $\vec{\epsilon} \sim \mathcal{N}(0, \sigma^2)$
is a Gaussian noise term
and the data-generating process
is the same of \Cref{subsec:dgp}.
Due to the presence of noise,
minimizing the least-squares error
does not ensure to recover
the true abstraction function.
We thus propose two strategies
to identify the concrete blocks
of each abstract variable.
By exploiting the fact
that each concrete variable
pertains to a single abstract variable,
we can filter the resulting matrix $\hat{\mat{T}}$
to select only the largest component per row
if it is above the threshold.
We find then beneficial
to refit the model once we have identified
the block in this way,
as in
\begin{align}
  \vec{t}_i = \argmin_{\vec{t}_i} \left\|
    \vec{x}_{\Pi_R{Y_i}} - \vec{t}_i^\tr \vec{y}_i
  \right\|^2_2.
\end{align}
In \Cref{fig:noisy},
we report results
for the reconstruction
of the blocks from the paired samples
for increasing
variance~$\sigma^2$
of the noise term
for these strategies.

\section{Additional Results}\label{app:additional}

\begin{figure}[H]
\centering
\begin{subfigure}[b]{0.32\textwidth}
         \centering
         \includegraphics[width=\textwidth]{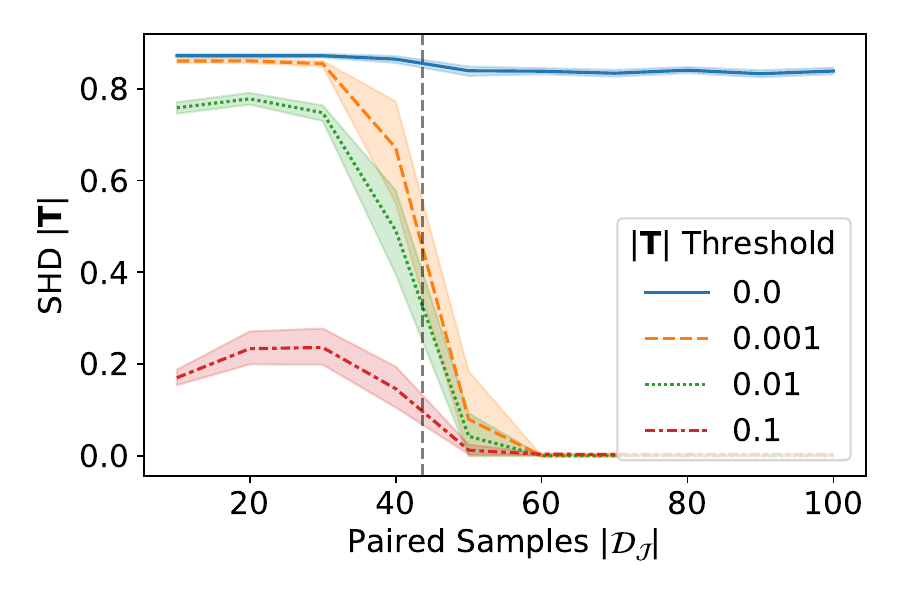}
         \caption{NHD}
     \end{subfigure}
     \begin{subfigure}[b]{0.32\textwidth}
         \centering
         \includegraphics[width=\textwidth]{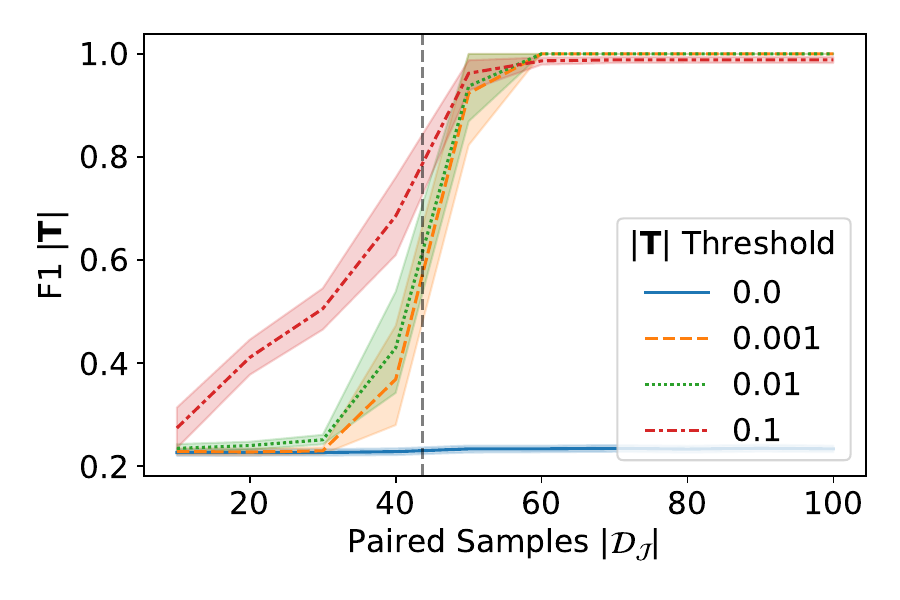}
         \caption{F1}
     \end{subfigure}
     \begin{subfigure}[b]{0.32\textwidth}
         \centering
         \includegraphics[width=\textwidth]{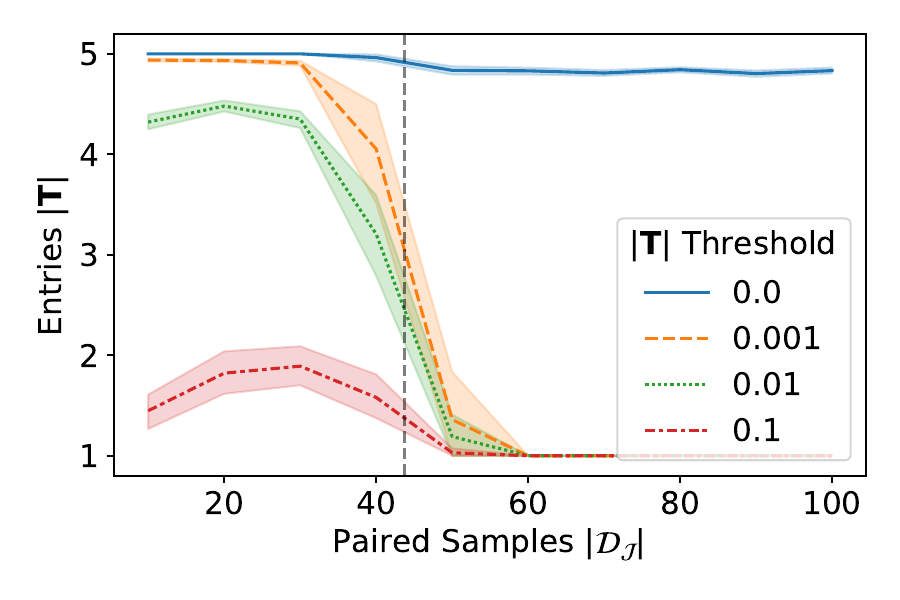}
         \caption{Abstract per Concrete}
     \end{subfigure}
\caption{%
    Reconstruction metrics
    of the linear abstraction function $\mathbf{T}$
    over pairs of abstract ($b=5$ nodes)
    and concrete ($d\in[25,50]$ nodes)
    linear SCMs
    for an increasing number
    of paired samples $|\mathcal{D_J}|$.
    For different thresholds,
    we report the normalized Hamming Distance (\emph{left}),
    the F1 score (\emph{center}),
    and the average number of abstract variables
    assigned to each concrete variable (\emph{right}).
    All results are averaged over 30 independent runs.
  }\label{fig:rec_t_small}
\end{figure}

\begin{figure}[H]
\centering
\begin{subfigure}[b]{0.32\textwidth}
         \centering
         \includegraphics[width=\textwidth]{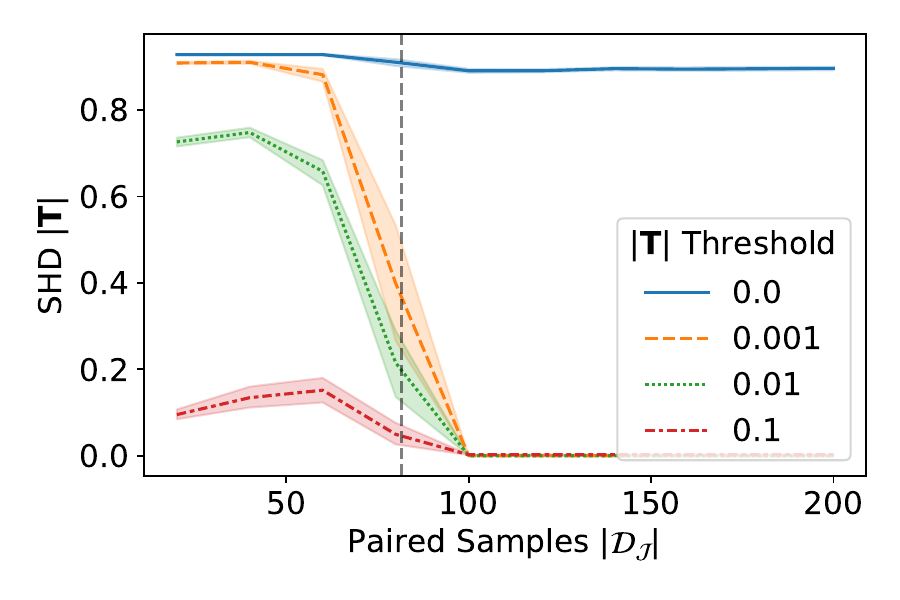}
         \caption{NHD}
     \end{subfigure}
     \begin{subfigure}[b]{0.32\textwidth}
         \centering
         \includegraphics[width=\textwidth]{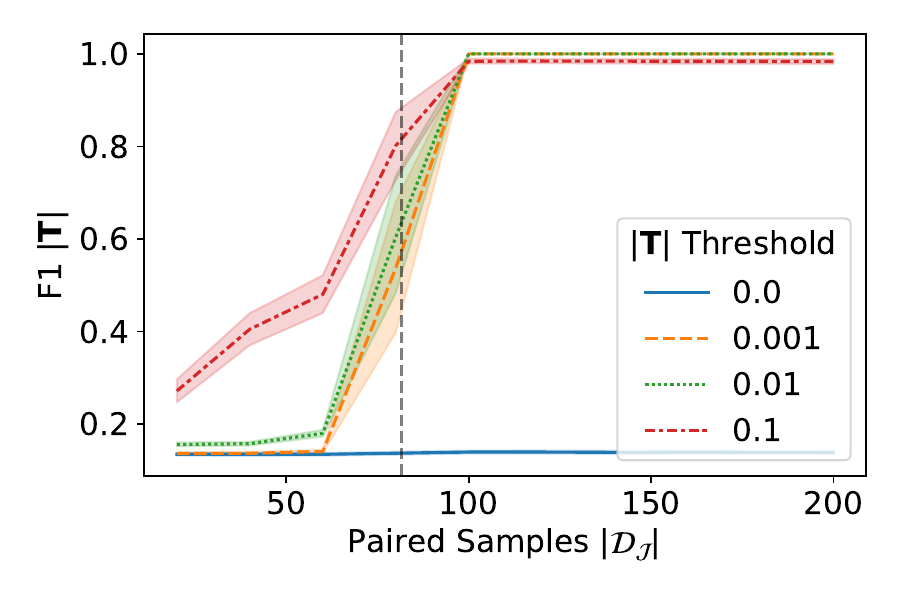}
         \caption{F1}
     \end{subfigure}
     \begin{subfigure}[b]{0.32\textwidth}
         \centering
         \includegraphics[width=\textwidth]{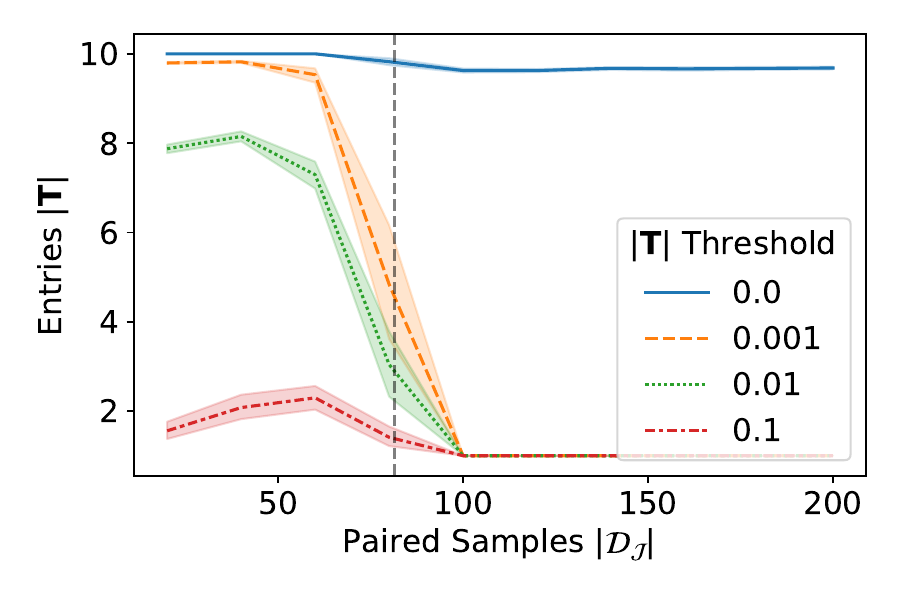}
         \caption{Abstract per Concrete}
     \end{subfigure}
\caption{%
    Reconstruction metrics
    of the linear abstraction function $\mathbf{T}$
    over pairs of abstract ($b=10$ nodes)
    and concrete ($d\in[50,100]$ nodes)
    linear SCMs
    for an increasing number
    of paired samples $|\mathcal{D_J}|$.
    For different thresholds,
    we report the normalized Hamming Distance (\emph{left}),
    the F1 score (\emph{center}),
    and the average number of abstract variables
    assigned to each concrete variable (\emph{right}).
    All results are averaged over 30 independent runs.
  }\label{fig:rec_t_medium}
\end{figure}

\begin{figure}[H]
\centering
\begin{subfigure}[b]{0.32\textwidth}
         \centering
         \includegraphics[width=\textwidth]{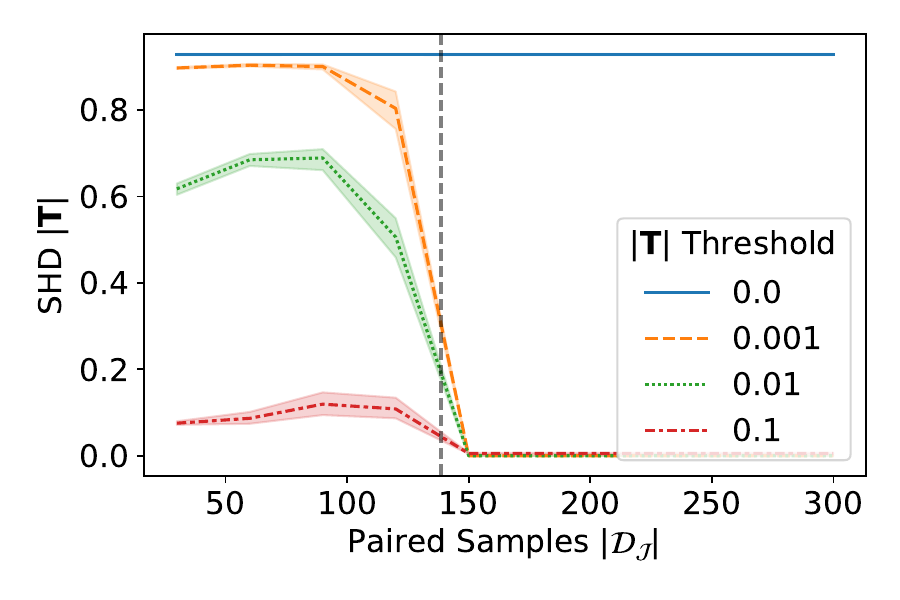}
         \caption{NHD}
     \end{subfigure}
     \begin{subfigure}[b]{0.32\textwidth}
         \centering
         \includegraphics[width=\textwidth]{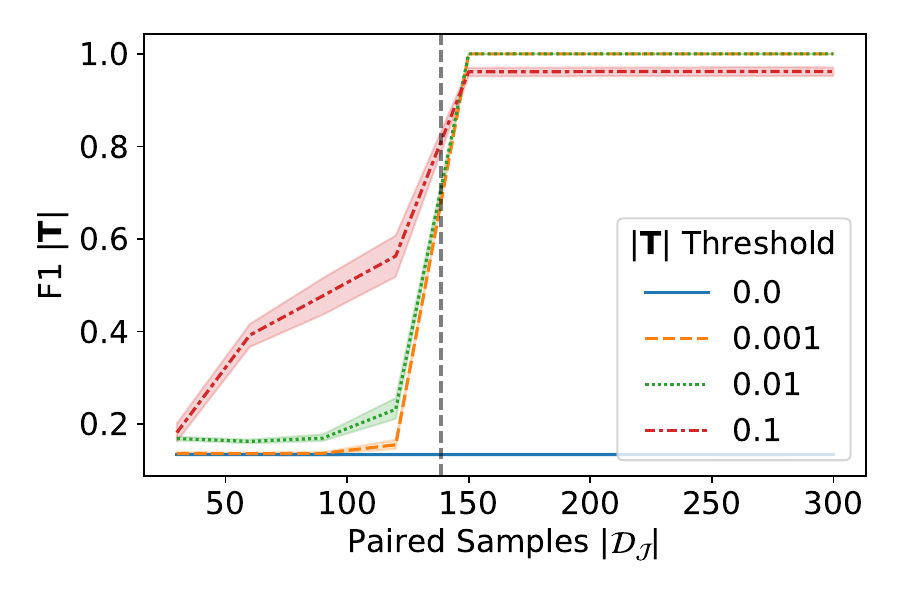}
         \caption{F1}
     \end{subfigure}
     \begin{subfigure}[b]{0.32\textwidth}
         \centering
         \includegraphics[width=\textwidth]{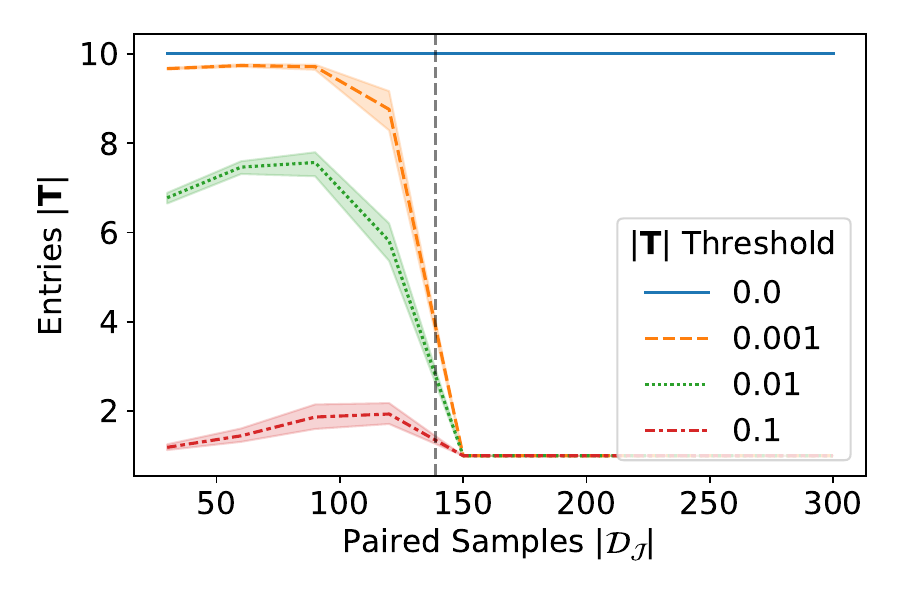}
         \caption{Abstract per Concrete}
     \end{subfigure}
\caption{%
    Reconstruction metrics
    of the linear abstraction function $\mathbf{T}$
    over pairs of abstract ($b=10$ nodes)
    and concrete ($d\in[100,150]$ nodes)
    linear SCMs
    for an increasing number
    of paired samples $|\mathcal{D_J}|$.
    For different thresholds,
    we report the normalized Hamming Distance (\emph{left}),
    the F1 score (\emph{center}),
    and the average number of abstract variables
    assigned to each concrete variable (\emph{right}).
    All results are averaged over 30 independent runs.
  }\label{fig:rec_t_large}
\end{figure}

In this section,
we report additional results
on our experiments
on Abs-LiNGAM (\Cref{alg:abslingam}).
We mostly consider three settings:
\emph{small}, where the number of nodes in the abstract model is $b=5$ and the number of nodes in the concrete model is $d\in[25,50]$; \emph{medium}, where the number of nodes in the abstract model is $b=10$ and the number of nodes in the concrete model is $d\in[50,100]$; and \emph{large}, where the number of nodes in the abstract model is $b=10$ and the number of nodes in the concrete model is $d\in[100,150]$.
We then report
results on the sensitivity of Abs-LiNGAM
to the number of paired samples $\dset_J$ (\Cref{fig:exp1_small,fig:exp1_medium,fig:exp1_large}),
the number of concrete samples $\dset_{\scm{L}}$ (\Cref{fig:exp2_small,fig:exp2_medium,fig:exp2_large}),
and the number of nodes in the concrete model $d$ (\Cref{fig:exp3_small,fig:exp3_medium,fig:exp3_large}).
Further, we report
results on the quality of the retrieved prior knowledge
given the threshold
used to mask the learned abstraction function~$\hat{\mat{T}}$ (\Cref{fig:exp5})
and the threshold used to mask the learned abstract model~$\hat{\scm{H}}$ (\Cref{fig:exp6}).
Similarly,
we study the retrieval of the prior knowledge
for different number of bootstrap samples
to identify the abstract model~$\hat{\scm{H}}$ (\Cref{fig:exp7}).
To provide further insights
on the performance of Abs-LiNGAM,
we also report precision and recall
on the three settings
(\Cref{table:small_aucprerec,table:medium_aucprerec,table:large_aucprerec}).
We finally report additional results on the reconstruction
of the abstraction function~$\hat{\mat{T}}$
in the small (\Cref{fig:rec_t_small}),
medium (\Cref{fig:rec_t_medium}),
and large (\Cref{fig:rec_t_large}) settings.

\onecolumn

\newcommand{\subfigwidth}{0.40\textwidth}
\newcommand{\subplotwidth}{0.5\textwidth}

\begin{figure}[H]
  \centering
  \begin{subfigure}[b]{\subfigwidth}
    \resizebox{\textwidth}{!}{\input{additional/exp1_concrete_roc_auc_small.pgf}}
  \end{subfigure}
  \begin{subfigure}[b]{\subfigwidth}
    \resizebox{\columnwidth}{!}{\input{additional/exp1_time_small.pgf}}
  \end{subfigure}\\
  \begin{subfigure}[b]{\subfigwidth}
    \resizebox{\textwidth}{!}{\input{additional/exp1_pk_precision_small.pgf}}
  \end{subfigure}
  \begin{subfigure}[b]{\subfigwidth}
    \resizebox{\textwidth}{!}{\input{additional/exp1_pk_recall_small.pgf}}
  \end{subfigure}
  \caption{
    Results of Abs-LiNGAM
    over pairs of abstract ($b=5$ nodes)
    and concrete ($d\in[25,50]$ nodes)
    linear SCMs.
    In all subfigures
    we plot the results
    for an increasing number of paired samples $\dset_J$
    and
    we report the average
    size of the concrete graphs
    as a vertical dashed line.
    Abs-LiNGAM-GT denotes
    a ground truth oracle
    where the abstraction function
    and the abstract model
    are given.
    The first plot (top left) shows the ROC-AUC
    of the retrieved concrete causal model $\hat{\scm{L}}$.
    The second plot (top right)
    shows the execution time
    required to retrieve the concrete causal model.
    The third and fourth plots (bottom) show the precision and recall of the prior knowledge
    inferred by the learned abstraction function~$\hat{\mat{T}}$
    and the consequent abstract model~$\hat{\scm{H}}$.
    All results are averaged over 30 independent runs
    with $|\dset_{\scm{L}}| = 15000$ concrete samples.
  }\label{fig:exp1_small}
\end{figure}

\begin{figure}[H]
      \centering
  \begin{subfigure}[b]{\subfigwidth}
    \resizebox{\textwidth}{!}{\input{additional/exp1_concrete_roc_auc_medium.pgf}}
  \end{subfigure}
  \begin{subfigure}[b]{\subfigwidth}
    \resizebox{\columnwidth}{!}{\input{additional/exp1_time_medium.pgf}}
  \end{subfigure}\\
  \begin{subfigure}[b]{\subfigwidth}
    \resizebox{\textwidth}{!}{\input{additional/exp1_pk_precision_medium.pgf}}
  \end{subfigure}
  \begin{subfigure}[b]{\subfigwidth}
    \resizebox{\textwidth}{!}{\input{additional/exp1_pk_recall_medium.pgf}}
  \end{subfigure}
  \caption{
    Results of Abs-LiNGAM
    over pairs of abstract ($b=10$ nodes)
    and concrete ($d\in[50,100]$ nodes)
    linear SCMs.
    In all subfigures
    we plot the results
    for an increasing number of paired samples $\dset_J$
    and
    we report the average
    size of the concrete graphs
    as a vertical dashed line.
    Abs-LiNGAM-GT denotes
    a ground truth oracle
    where the abstraction function
    and the abstract model
    are given.
    The first plot (top left) shows the ROC-AUC
    of the retrieved concrete causal model $\hat{\scm{L}}$.
    The second plot (top right)
    shows the execution time
    required to retrieve the concrete causal model.
    The third and fourth plots (bottom) show the precision and recall of the prior knowledge
    inferred by the learned abstraction function~$\hat{\mat{T}}$
    and the consequent abstract model~$\hat{\scm{H}}$.
    All results are averaged over 30 independent runs
    with $|\dset_{\scm{L}}| = 15000$ concrete samples.
  }\label{fig:exp1_medium}
\end{figure}

\begin{figure}[H]
      \centering
  \begin{subfigure}[b]{\subfigwidth}
    \resizebox{\textwidth}{!}{\input{additional/exp1_concrete_roc_auc_large.pgf}}
  \end{subfigure}
  \begin{subfigure}[b]{\subfigwidth}
    \resizebox{\columnwidth}{!}{\input{additional/exp1_time_large.pgf}}
  \end{subfigure}\\
  \begin{subfigure}[b]{\subfigwidth}
    \resizebox{\textwidth}{!}{\input{additional/exp1_pk_precision_large.pgf}}
  \end{subfigure}
  \begin{subfigure}[b]{\subfigwidth}
    \resizebox{\textwidth}{!}{\input{additional/exp1_pk_recall_large.pgf}}
  \end{subfigure}
  \caption{
    Results of Abs-LiNGAM
    over pairs of abstract ($b=10$ nodes)
    and concrete ($d\in[100,150]$ nodes)
    linear SCMs.
    In all subfigures
    we plot the results
    for an increasing number of paired samples $\dset_J$
    and
    we report the average
    size of the concrete graphs
    as a vertical dashed line.
    Abs-LiNGAM-GT denotes
    a ground truth oracle
    where the abstraction function
    and the abstract model
    are given.
    The first plot (top left) shows the ROC-AUC
    of the retrieved concrete causal model $\hat{\scm{L}}$.
    The second plot (top right)
    shows the execution time
    required to retrieve the concrete causal model.
    The third and fourth plots (bottom) show the precision and recall of the prior knowledge
    inferred by the learned abstraction function~$\hat{\mat{T}}$
    and the consequent abstract model~$\hat{\scm{H}}$.
    All results are averaged over 30 independent runs
    with $|\dset_{\scm{L}}| = 15000$ concrete samples.
  }\label{fig:exp1_large}
\end{figure}

\begin{figure}[H]
      \centering
  \begin{subfigure}[b]{\subfigwidth}
    \resizebox{\textwidth}{!}{\input{additional/exp2_concrete_roc_auc_small.pgf}}
  \end{subfigure}
  \begin{subfigure}[b]{\subfigwidth}
    \resizebox{\columnwidth}{!}{\input{additional/exp2_time_small.pgf}}
  \end{subfigure}\\
  \begin{subfigure}[b]{\subfigwidth}
    \resizebox{\textwidth}{!}{\input{additional/exp2_pk_precision_small.pgf}}
  \end{subfigure}
  \begin{subfigure}[b]{\subfigwidth}
    \resizebox{\textwidth}{!}{\input{additional/exp2_pk_recall_small.pgf}}
  \end{subfigure}
  \caption{%
    Results of Abs-LiNGAM
    over pairs of abstract ($b=5$ nodes)
    and concrete ($d\in[25,50]$ nodes)
    linear SCMs.
    In all subfigures
    we plot the results
    for an increasing number
    of concrete samples $|\dset_{\scm{L}}|$.
    Abs-LiNGAM-GT denotes
    a ground truth oracle
    where the abstraction function
    and the abstract model
    are given.
    The first plot (top left) shows the ROC-AUC
    of the retrieved concrete causal model $\hat{\scm{L}}$.
    The second plot (top right)
    shows the execution time
    required to retrieve the concrete causal model.
    The third and fourth plots (bottom) show the precision and recall of the prior knowledge
    inferred by the learned abstraction function~$\hat{\mat{T}}$
    and the consequent abstract model~$\hat{\scm{H}}$.
    All results are averaged over 30 independent runs
    with $|\dset_J| = 100$ paired samples.
  }\label{fig:exp2_small}
\end{figure}

\begin{figure}[H]
      \centering
  \begin{subfigure}[b]{\subfigwidth}
    \resizebox{\textwidth}{!}{\input{additional/exp2_concrete_roc_auc_medium.pgf}}
  \end{subfigure}
  \begin{subfigure}[b]{\subfigwidth}
    \resizebox{\columnwidth}{!}{\input{additional/exp2_time_medium.pgf}}
  \end{subfigure}\\
  \begin{subfigure}[b]{\subfigwidth}
    \resizebox{\textwidth}{!}{\input{additional/exp2_pk_precision_medium.pgf}}
  \end{subfigure}
  \begin{subfigure}[b]{\subfigwidth}
    \resizebox{\textwidth}{!}{\input{additional/exp2_pk_recall_medium.pgf}}
  \end{subfigure}
  \caption{%
    Results of Abs-LiNGAM
    over pairs of abstract ($b=10$ nodes)
    and concrete ($d\in[50,100]$ nodes)
    linear SCMs.
    In all subfigures
    we plot the results
    for an increasing number
    of concrete samples $|\dset_{\scm{L}}|$.
    Abs-LiNGAM-GT denotes
    a ground truth oracle
    where the abstraction function
    and the abstract model
    are given.
    The first plot (top left) shows the ROC-AUC
    of the retrieved concrete causal model $\hat{\scm{L}}$.
    The second plot (top right)
    shows the execution time
    required to retrieve the concrete causal model.
    The third and fourth plots (bottom) show the precision and recall of the prior knowledge
    inferred by the learned abstraction function~$\hat{\mat{T}}$
    and the consequent abstract model~$\hat{\scm{H}}$.
    All results are averaged over 30 independent runs
    with $|\dset_J| = 200$ paired samples.
  }\label{fig:exp2_medium}
\end{figure}

\begin{figure}[H]
      \centering
  \begin{subfigure}[b]{\subfigwidth}
    \resizebox{\textwidth}{!}{\input{additional/exp2_concrete_roc_auc_large.pgf}}
  \end{subfigure}
  \begin{subfigure}[b]{\subfigwidth}
    \resizebox{\columnwidth}{!}{\input{additional/exp2_time_large.pgf}}
  \end{subfigure}\\
  \begin{subfigure}[b]{\subfigwidth}
    \resizebox{\textwidth}{!}{\input{additional/exp2_pk_precision_large.pgf}}
  \end{subfigure}
  \begin{subfigure}[b]{\subfigwidth}
    \resizebox{\textwidth}{!}{\input{additional/exp2_pk_recall_large.pgf}}
  \end{subfigure}
  \caption{%
    Results of Abs-LiNGAM
    over pairs of abstract ($b=10$ nodes)
    and concrete ($d\in[100,150]$ nodes)
    linear SCMs.
    In all subfigures
    we plot the results
    for an increasing number
    of concrete samples $|\dset_{\scm{L}}|$.
    Abs-LiNGAM-GT denotes
    a ground truth oracle
    where the abstraction function
    and the abstract model
    are given.
    The first plot (top left) shows the ROC-AUC
    of the retrieved concrete causal model $\hat{\scm{L}}$.
    The second plot (top right)
    shows the execution time
    required to retrieve the concrete causal model.
    The third and fourth plots (bottom) show the precision and recall of the prior knowledge
    inferred by the learned abstraction function~$\hat{\mat{T}}$
    and the consequent abstract model~$\hat{\scm{H}}$.
    All results are averaged over 30 independent runs
    with $|\dset_J| = 300$ paired samples.
  }\label{fig:exp2_large}
\end{figure}

\begin{figure}[H]
      \centering
  \begin{subfigure}[b]{\subfigwidth}
    \resizebox{\textwidth}{!}{\input{additional/exp3_concrete_roc_auc_small.pgf}}
  \end{subfigure}
  \begin{subfigure}[b]{\subfigwidth}
    \resizebox{\columnwidth}{!}{\input{additional/exp3_time_small.pgf}}
  \end{subfigure}\\
  \begin{subfigure}[b]{\subfigwidth}
    \resizebox{\textwidth}{!}{\input{additional/exp3_pk_precision_small.pgf}}
  \end{subfigure}
  \begin{subfigure}[b]{\subfigwidth}
    \resizebox{\textwidth}{!}{\input{additional/exp3_pk_recall_small.pgf}}
  \end{subfigure}
  \caption{%
    Results of Abs-LiNGAM
    over pairs of abstract ($b=5$ nodes)
    and concrete models with increasing size $d\in[5,60]$.
    Abs-LiNGAM-GT denotes
    a ground truth oracle
    where the abstraction function
    and the abstract model
    are given.
    The first plot (top left) shows the ROC-AUC
    of the retrieved concrete causal model $\hat{\scm{L}}$.
    The second plot (top right)
    shows the execution time
    required to retrieve the concrete causal model.
    The third and fourth plots (bottom) show the precision and recall of the prior knowledge
    inferred by the learned abstraction function~$\hat{\mat{T}}$
    and the consequent abstract model~$\hat{\scm{H}}$.
    All results are averaged over 30 independent runs
    with $|\dset_{\scm{L}}| = 1500$ concrete samples
    and $|dset_J| = 2 \cdot |\set{X}|$ paired samples.
  }\label{fig:exp3_small}
\end{figure}

\begin{figure}[H]
      \centering
  \begin{subfigure}[b]{\subfigwidth}
    \resizebox{\textwidth}{!}{\input{additional/exp3_concrete_roc_auc_medium.pgf}}
  \end{subfigure}
  \begin{subfigure}[b]{\subfigwidth}
    \resizebox{\columnwidth}{!}{\input{additional/exp3_time_medium.pgf}}
  \end{subfigure}\\
  \begin{subfigure}[b]{\subfigwidth}
    \resizebox{\textwidth}{!}{\input{additional/exp3_pk_precision_medium.pgf}}
  \end{subfigure}
  \begin{subfigure}[b]{\subfigwidth}
    \resizebox{\textwidth}{!}{\input{additional/exp3_pk_recall_medium.pgf}}
  \end{subfigure}
  \caption{%
    Results of Abs-LiNGAM
    over pairs of abstract ($b=10$ nodes)
    and concrete models with increasing size $d\in[10,120]$.
    Abs-LiNGAM-GT denotes
    a ground truth oracle
    where the abstraction function
    and the abstract model
    are given.
    The first plot (top left) shows the ROC-AUC
    of the retrieved concrete causal model $\hat{\scm{L}}$.
    The second plot (top right)
    shows the execution time
    required to retrieve the concrete causal model.
    The third and fourth plots (bottom) show the precision and recall of the prior knowledge
    inferred by the learned abstraction function~$\hat{\mat{T}}$
    and the consequent abstract model~$\hat{\scm{H}}$.
    All results are averaged over 30 independent runs
    with $|\dset_{\scm{L}}| = 1500$ concrete samples
    and $|dset_J| = 2 \cdot |\set{X}|$ paired samples.
  }\label{fig:exp3_medium}
\end{figure}

\begin{figure}[H]
      \centering
  \begin{subfigure}[b]{\subfigwidth}
    \resizebox{\textwidth}{!}{\input{additional/exp3_concrete_roc_auc_large.pgf}}
  \end{subfigure}
  \begin{subfigure}[b]{\subfigwidth}
    \resizebox{\columnwidth}{!}{\input{additional/exp3_time_large.pgf}}
  \end{subfigure}\\
  \begin{subfigure}[b]{\subfigwidth}
    \resizebox{\textwidth}{!}{\input{additional/exp3_pk_precision_large.pgf}}
  \end{subfigure}
  \begin{subfigure}[b]{\subfigwidth}
    \resizebox{\textwidth}{!}{\input{additional/exp3_pk_recall_large.pgf}}
  \end{subfigure}
  \caption{%
    Results of Abs-LiNGAM
    over pairs of abstract ($b=10$ nodes)
    and concrete models with increasing size $d\in[20,180]$.
    Abs-LiNGAM-GT denotes
    a ground truth oracle
    where the abstraction function
    and the abstract model
    are given.
    The first plot (top left) shows the ROC-AUC
    of the retrieved concrete causal model $\hat{\scm{L}}$.
    The second plot (top right)
    shows the execution time
    required to retrieve the concrete causal model.
    The third and fourth plots (bottom) show the precision and recall of the prior knowledge
    inferred by the learned abstraction function~$\hat{\mat{T}}$
    and the consequent abstract model~$\hat{\scm{H}}$.
    All results are averaged over 30 independent runs
    with $|\dset_{\scm{L}}| = 1500$ concrete samples
    and $|dset_J| = 2 \cdot |\set{X}|$ paired samples.
  }\label{fig:exp3_large}
\end{figure}

\begin{figure}[H]
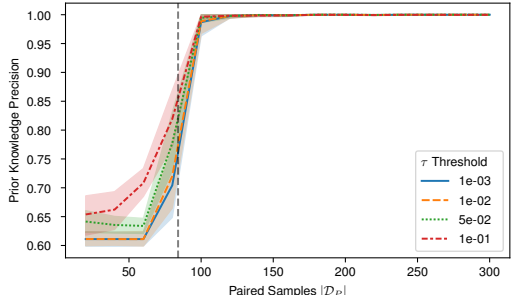
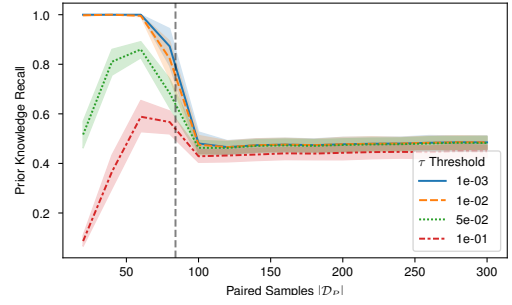
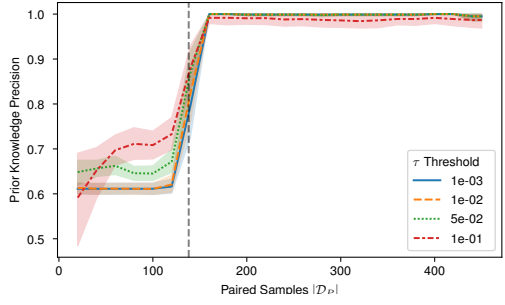
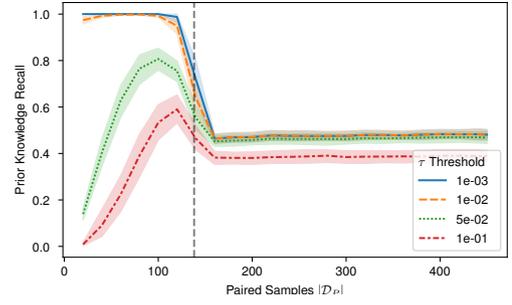

  \centering
  \begin{minipage}{0.11\textwidth}
    \centering
    \small{
      \begin{align*}
        b &= 5 \\
        d &\in [25,50]
    \end{align*}}
  \end{minipage}
  \hfill
  \begin{minipage}{0.88\textwidth}
    \begin{subfigure}[b]{\subplotwidth}
      \resizebox{\columnwidth}{!}{\input{additional/exp5_pk_precision_small.pgf}}
    \end{subfigure}
    \begin{subfigure}[b]{\subplotwidth}
      \resizebox{\columnwidth}{!}{\input{additional/exp5_pk_recall_small.pgf}}
    \end{subfigure}
  \end{minipage}\\
  \begin{minipage}{0.11\textwidth}
    \centering
    \small{
      \begin{align*}
        b &= 10 \\
        d &\in [50,100]
    \end{align*}}
  \end{minipage}
  \hfill
  \begin{minipage}{0.88\textwidth}
    \begin{subfigure}[b]{\subplotwidth}
      \resizebox{\columnwidth}{!}{\input{additional/exp5_pk_precision_medium.pgf}}
    \end{subfigure}
    \begin{subfigure}[b]{\subplotwidth}
      \resizebox{\columnwidth}{!}{\input{additional/exp5_pk_recall_medium.pgf}}
    \end{subfigure}
  \end{minipage}\\
  \begin{minipage}{0.11\textwidth}
    \centering
    \small{
      \begin{align*}
        b &= 10 \\
        d &\in [100,150]
    \end{align*}}
  \end{minipage}
  \hfill
  \begin{minipage}{0.88\textwidth}
    \begin{subfigure}[b]{\subplotwidth}
      \resizebox{\columnwidth}{!}{\input{additional/exp5_pk_precision_large.pgf}}
    \end{subfigure}
    \begin{subfigure}[b]{\subplotwidth}
      \resizebox{\columnwidth}{!}{\input{additional/exp5_pk_recall_large.pgf}}
    \end{subfigure}
  \end{minipage}
  \caption{%
    Analysis of the prior knowledge
    inferred by the learned abstraction function~$\hat{\mat{T}}$
    and the consequent abstract model~$\hat{\scm{H}}$
    on a concrete model ($d\in[25,50]$ nodes).
    We report precision (left) and recall (right)
    of the prior knowledge
    for different thresholds
    to mask the learned abstraction function~$\hat{\mat{T}}$.
  }\label{fig:exp5}
\end{figure}

\begin{figure}[H]
  \centering
  \begin{minipage}{0.11\textwidth}
    \centering
    \small{
      \begin{align*}
        b &= 5 \\
        d &\in [25,50]
    \end{align*}}
  \end{minipage}
  \hfill
  \begin{minipage}{0.88\textwidth}
    \begin{subfigure}[b]{\subplotwidth}
      \resizebox{\columnwidth}{!}{\input{additional/exp6_pk_precision_small.pgf}}
    \end{subfigure}
    \begin{subfigure}[b]{\subplotwidth}
      \resizebox{\columnwidth}{!}{\input{additional/exp6_pk_recall_small.pgf}}
    \end{subfigure}
  \end{minipage}\\
  \begin{minipage}{0.11\textwidth}
    \centering
    \small{
      \begin{align*}
        b &= 10 \\
        d &\in [50,100]
    \end{align*}}
  \end{minipage}
  \hfill
  \begin{minipage}{0.88\textwidth}
    \begin{subfigure}[b]{\subplotwidth}
      \resizebox{\columnwidth}{!}{\input{additional/exp6_pk_precision_medium.pgf}}
    \end{subfigure}
    \begin{subfigure}[b]{\subplotwidth}
      \resizebox{\columnwidth}{!}{\input{additional/exp6_pk_recall_medium.pgf}}
    \end{subfigure}
  \end{minipage}\\
  \begin{minipage}{0.11\textwidth}
    \centering
    \small{
      \begin{align*}
        b &= 10 \\
        d &\in [100,150]
    \end{align*}}
  \end{minipage}
  \hfill
  \begin{minipage}{0.88\textwidth}
    \begin{subfigure}[b]{\subplotwidth}
      \resizebox{\columnwidth}{!}{\input{additional/exp6_pk_precision_large.pgf}}
    \end{subfigure}
    \begin{subfigure}[b]{\subplotwidth}
      \resizebox{\columnwidth}{!}{\input{additional/exp6_pk_recall_large.pgf}}
    \end{subfigure}
  \end{minipage}
  \caption{%
    Analysis of the prior knowledge
    inferred by the learned abstraction function~$\hat{\mat{T}}$
    and the consequent abstract model~$\hat{\scm{H}}$
    on a concrete model.
    We report precision (left) and recall (right)
    of the prior knowledge
    for different thresholds
    to mask the learned abstract model~$\hat{\scm{H}}$.
  }\label{fig:exp6}
\end{figure}

\begin{figure}[H]
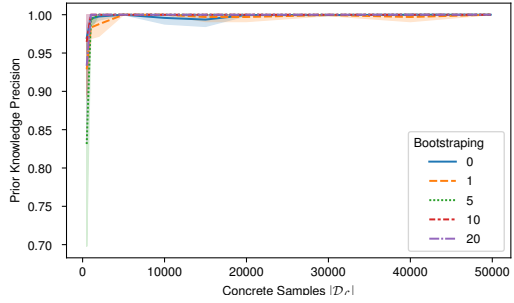
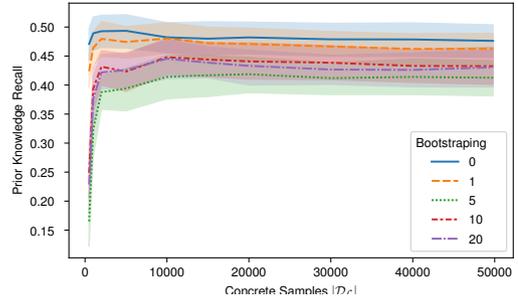
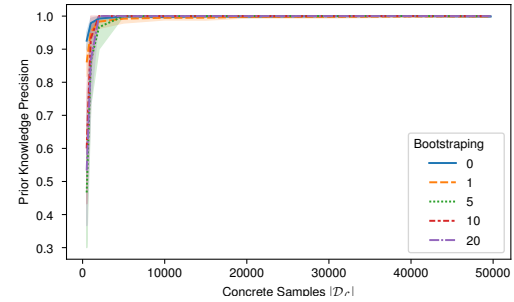
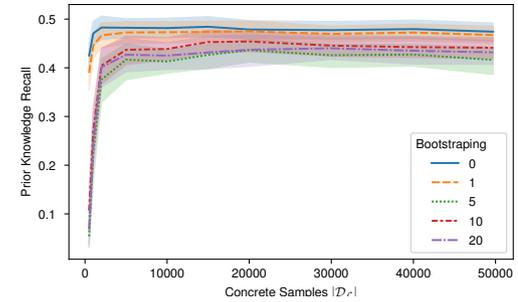

  \centering
  \begin{minipage}{0.11\textwidth}
    \centering
    \small{
      \begin{align*}
        b &= 5 \\
        d &\in [25,50]
    \end{align*}}
  \end{minipage}
  \hfill
  \begin{minipage}{0.88\textwidth}
    \begin{subfigure}[b]{\subplotwidth}
      \resizebox{\columnwidth}{!}{\input{additional/exp7_pk_precision_small.pgf}}
    \end{subfigure}
    \begin{subfigure}[b]{\subplotwidth}
      \resizebox{\columnwidth}{!}{\input{additional/exp7_pk_recall_small.pgf}}
    \end{subfigure}
  \end{minipage}\\
  \begin{minipage}{0.11\textwidth}
    \centering
    \small{
      \begin{align*}
        b &= 5 \\
        d &\in [25,50]
    \end{align*}}
  \end{minipage}
  \hfill
  \begin{minipage}{0.88\textwidth}
    \begin{subfigure}[b]{\subplotwidth}
      \resizebox{\columnwidth}{!}{\input{additional/exp7_pk_precision_medium.pgf}}
    \end{subfigure}
    \begin{subfigure}[b]{\subplotwidth}
      \resizebox{\columnwidth}{!}{\input{additional/exp7_pk_recall_medium.pgf}}
    \end{subfigure}
  \end{minipage}\\
  \begin{minipage}{0.11\textwidth}
    \centering
    \small{
      \begin{align*}
        b &= 5 \\
        d &\in [25,50]
    \end{align*}}
  \end{minipage}
  \hfill
  \begin{minipage}{0.88\textwidth}
    \begin{subfigure}[b]{\subplotwidth}
      \resizebox{\columnwidth}{!}{\input{additional/exp7_pk_precision_large.pgf}}
    \end{subfigure}
    \begin{subfigure}[b]{\subplotwidth}
      \resizebox{\columnwidth}{!}{\input{additional/exp7_pk_recall_large.pgf}}
    \end{subfigure}
  \end{minipage}
  \caption{%
    Analysis of the prior knowledge
    inferred by the learned abstraction function~$\hat{\mat{T}}$
    and the consequent abstract model~$\hat{\scm{H}}$,
    with $b$ nodes,
    on a concrete model with $d$ nodes.
    We report precision (left) and recall (right)
    of the prior knowledge
    for different number
    of bootstrapped samples
    to fit the abstract model~$\hat{\scm{H}}$.
  }\label{fig:exp7}
\end{figure}

\newpage
\begin{table}[H]
\centering
\begin{tabular}{lllll}
Method                    & ROCAUC    & Precision & Recall    & Time    \\
\midrule
Abs-Fit (Bootstrap=0)     &0.965$\pm$0.066&0.957$\pm$0.123&0.940$\pm$0.074&42$\pm$13 \\
Abs-Fit (Bootstrap=1)     &0.977$\pm$0.012&0.980$\pm$0.015&0.953$\pm$0.026&42$\pm$14 \\
Abs-Fit (Bootstrap=2)     &0.977$\pm$0.012&0.980$\pm$0.015&0.952$\pm$0.027&43$\pm$11 \\
Abs-Fit (Bootstrap=5)     &0.977$\pm$0.012&0.980$\pm$0.015&0.952$\pm$0.027&45$\pm$13 \\
Abs-Fit (Bootstrap=10)    &0.977$\pm$0.012&0.980$\pm$0.015&0.952$\pm$0.027&46$\pm$12 \\
Abs-LiNGAM-GT             &0.977$\pm$0.011&0.982$\pm$0.013&0.953$\pm$0.026&45$\pm$15 \\
DirectLiNGAM              &0.977$\pm$0.011&0.980$\pm$0.013&0.953$\pm$0.026&61$\pm$12 \\
\bottomrule
\end{tabular}
\caption{%
    Results of Abs-LiNGAM
    over pairs of abstract ($b=5$ nodes)
    and concrete ($d\in[25,50]$ nodes)
    linear SCMs.
    Abs-LiNGAM-GT denotes
    a ground truth oracle
    where the abstraction function
    and the abstract model
    are given.
    All results are averaged over 30 independent runs
    with $|\mathcal{D}_{\mathcal{L}}| = 15000$
    concrete
    and
    $|\mathcal{D}_J| = 150$
    paired samples.
  }\label{table:small_aucprerec}
\end{table}

\begin{table}[H]
\centering
\begin{tabular}{lllll}
Method                    & ROCAUC    & Precision & Recall    & Time    \\
\midrule
Abs-LiNGAM (Bootstrap=0)  &0.963$\pm$0.043&0.939$\pm$0.119&0.926$\pm$0.067&179$\pm$53\\
Abs-LiNGAM (Bootstrap=1)  &0.952$\pm$0.066&0.914$\pm$0.169&0.914$\pm$0.085&181$\pm$53\\
Abs-LiNGAM (Bootstrap=2)  &0.968$\pm$0.027&0.956$\pm$0.041&0.930$\pm$0.061&182$\pm$50\\
Abs-LiNGAM (Bootstrap=5)  &0.968$\pm$0.027&0.955$\pm$0.041&0.930$\pm$0.061&189$\pm$51\\
Abs-LiNGAM (Bootstrap=10) &0.968$\pm$0.027&0.954$\pm$0.040&0.930$\pm$0.061&194$\pm$51\\
Abs-LiNGAM-GT             &0.969$\pm$0.026&0.965$\pm$0.022&0.931$\pm$0.060&186$\pm$54\\
DirectLiNGAM              &0.968$\pm$0.025&0.958$\pm$0.020&0.930$\pm$0.061&394$\pm$94\\
\bottomrule
\end{tabular}
\caption{%
    Results of Abs-LiNGAM
    over pairs of abstract ($b=10$ nodes)
    and concrete ($d\in[50,100]$ nodes)
    linear SCMs.
    Abs-LiNGAM-GT denotes
    a ground truth oracle
    where the abstraction function
    and the abstract model
    are given.
    All results are averaged over 30 independent runs
    with $|\mathcal{D}_{\mathcal{L}}| = 15000$
    concrete
    and
    $|\mathcal{D}_J| = 270$
    paired samples.
}\label{table:medium_aucprerec}
\end{table}

\begin{table}[H]
\centering
\begin{tabular}{lllll}
Method                    & ROCAUC    & Precision & Recall    & Time    \\
\midrule
Abs-LiNGAM                &0.927$\pm$0.070&0.919$\pm$0.119&0.845$\pm$0.132&748$\pm$121  \\
Abs-LiNGAM (Bootstrap=1)  &0.913$\pm$0.083&0.877$\pm$0.187&0.834$\pm$0.136&731$\pm$116  \\
Abs-LiNGAM (Bootstrap=2)  &0.925$\pm$0.072&0.912$\pm$0.130&0.844$\pm$0.132&738$\pm$123  \\
Abs-LiNGAM (Bootstrap=5)  &0.926$\pm$0.067&0.913$\pm$0.109&0.844$\pm$0.130&755$\pm$140  \\
Abs-LiNGAM (Bootstrap=10) &0.927$\pm$0.065&0.918$\pm$0.090&0.844$\pm$0.130&775$\pm$183  \\
Abs-LiNGAM-GT             &0.927$\pm$0.069&0.920$\pm$0.117&0.845$\pm$0.131&763$\pm$116  \\
DirectLiNGAM              &0.928$\pm$0.061&0.925$\pm$0.047&0.844$\pm$0.128&1608$\pm$212 \\
\bottomrule
\end{tabular}
\caption{%
    Results of Abs-LiNGAM
    over pairs of abstract ($b=10$ nodes)
    and concrete ($d\in[100,150]$ nodes)
    linear SCMs.
    Abs-LiNGAM-GT denotes
    a ground truth oracle
    where the abstraction function
    and the abstract model
    are given.
    All results are averaged over 30 independent runs
    with $|\mathcal{D}_{\mathcal{L}}| = 15000$
    concrete
    and
    $|\mathcal{D}_J| = 270$
    paired samples.
}\label{table:large_aucprerec}
\end{table}

\newpage

\begin{thebibliography}{24}
\providecommand{\natexlab}[1]{#1}
\providecommand{\url}[1]{\texttt{#1}}
\expandafter\ifx\csname urlstyle\endcsname\relax
  \providecommand{\doi}[1]{doi: #1}\else
  \providecommand{\doi}{doi: \begingroup \urlstyle{rm}\Url}\fi

\bibitem[Anand et~al.(2023)Anand, Ribeiro, Tian, and
  Bareinboim]{anand2023causal}
Tara~V Anand, Adele~H Ribeiro, Jin Tian, and Elias Bareinboim.
\newblock Causal effect identification in cluster dags.
\newblock In \emph{Proceedings of the AAAI Conference on Artificial
  Intelligence}, volume~37, pages 12172--12179, 2023.

\bibitem[Beckers and Halpern(2019)]{beckers2019abstracting}
Sander Beckers and Joseph~Y Halpern.
\newblock Abstracting causal models.
\newblock In \emph{Proceedings of the aaai conference on artificial
  intelligence}, volume~33, pages 2678--2685, 2019.

\bibitem[Bondy and Murty(2008)]{bondy_graph_2008}
Adrian Bondy and M.~Ram Murty.
\newblock \emph{Graph Theory}.
\newblock Graduate Texts in Mathematics. Springer-Verlag, 2008.
\newblock ISBN 978-1-84628-969-9.
\newblock URL \url{https://www.springer.com/gp/book/9781846289699}.

\bibitem[Bongers et~al.(2021)Bongers, Forr{\'e}, Peters, and
  Mooij]{bongers2021foundations}
Stephan Bongers, Patrick Forr{\'e}, Jonas Peters, and Joris~M Mooij.
\newblock Foundations of structural causal models with cycles and latent
  variables.
\newblock \emph{The Annals of Statistics}, 49\penalty0 (5):\penalty0
  2885--2915, 2021.

\bibitem[Chalupka et~al.(2016)Chalupka, Bischoff, Perona, and
  Eberhardt]{chalupka2016unsupervised}
Krzysztof Chalupka, Tobias Bischoff, Pietro Perona, and Frederick Eberhardt.
\newblock Unsupervised discovery of el nino using causal feature learning on
  microlevel climate data.
\newblock In \emph{Proceedings of the Thirty-Second Conference on Uncertainty
  in Artificial Intelligence}, pages 72--81, 2016.

\bibitem[Dubois et~al.(2020)Dubois, Eberhardt, Paul, and
  Adolphs]{dubois2020personality}
Julien Dubois, Frederick Eberhardt, Lynn~K Paul, and Ralph Adolphs.
\newblock Personality beyond taxonomy.
\newblock \emph{Nature human behaviour}, 4\penalty0 (11):\penalty0 1110--1117,
  2020.

\bibitem[Felekis et~al.(2024)Felekis, Zennaro, Branchini, and
  Damoulas]{felekis2024causal}
Yorgos Felekis, Fabio~Massimo Zennaro, Nicola Branchini, and Theodoros
  Damoulas.
\newblock Causal optimal transport of abstractions.
\newblock In \emph{Causal Learning and Reasoning}, pages 462--498. PMLR, 2024.

\bibitem[Geiger et~al.(2021)Geiger, Lu, Icard, and Potts]{geiger2021causal}
Atticus Geiger, Hanson Lu, Thomas Icard, and Christopher Potts.
\newblock Causal abstractions of neural networks.
\newblock \emph{Advances in Neural Information Processing Systems},
  34:\penalty0 9574--9586, 2021.

\bibitem[Geiger et~al.(2023)Geiger, Potts, and Icard]{geiger2023causal}
Atticus Geiger, Chris Potts, and Thomas Icard.
\newblock Causal abstraction for faithful model interpretation.
\newblock \emph{arXiv preprint arXiv:2301.04709}, 2023.

\bibitem[Geiger et~al.(2024)Geiger, Wu, Potts, Icard, and
  Goodman]{geiger2024finding}
Atticus Geiger, Zhengxuan Wu, Christopher Potts, Thomas Icard, and Noah
  Goodman.
\newblock Finding alignments between interpretable causal variables and
  distributed neural representations.
\newblock In \emph{Causal Learning and Reasoning}, pages 160--187. PMLR, 2024.

\bibitem[Keki{\'c} et~al.(2023)Keki{\'c}, Sch{\"o}lkopf, and
  Besserve]{kekic2023targeted}
Armin Keki{\'c}, Bernhard Sch{\"o}lkopf, and Michel Besserve.
\newblock Targeted reduction of causal models.
\newblock \emph{arXiv preprint arXiv:2311.18639}, 2023.

\bibitem[Marconato et~al.(2023)Marconato, Passerini, and
  Teso]{marconato2023interpretability}
Emanuele Marconato, Andrea Passerini, and Stefano Teso.
\newblock Interpretability is in the mind of the beholder: A causal framework
  for human-interpretable representation learning.
\newblock \emph{Entropy}, 25\penalty0 (12):\penalty0 1574, 2023.

\bibitem[Massidda et~al.(2023)Massidda, Geiger, Icard, and
  Bacciu]{massidda2023causal}
Riccardo Massidda, Atticus Geiger, Thomas Icard, and Davide Bacciu.
\newblock Causal abstraction with soft interventions.
\newblock In \emph{Conference on Causal Learning and Reasoning}, pages 68--87.
  PMLR, 2023.

\bibitem[Pearl(2009)]{pearl2009causality}
Judea Pearl.
\newblock \emph{Causality}.
\newblock Cambridge university press, 2009.

\bibitem[Peters et~al.(2017)Peters, Janzing, and
  Sch{\"o}lkopf]{peters2017elements}
Jonas Peters, Dominik Janzing, and Bernhard Sch{\"o}lkopf.
\newblock \emph{Elements of causal inference: foundations and learning
  algorithms}.
\newblock The MIT Press, 2017.

\bibitem[Shimizu et~al.(2006)Shimizu, Hoyer, Hyv{\"a}rinen, Kerminen, and
  Jordan]{shimizu2006linear}
Shohei Shimizu, Patrik~O Hoyer, Aapo Hyv{\"a}rinen, Antti Kerminen, and Michael
  Jordan.
\newblock A linear non-gaussian acyclic model for causal discovery.
\newblock \emph{Journal of Machine Learning Research}, 7\penalty0 (10), 2006.

\bibitem[Shimizu et~al.(2011)Shimizu, Inazumi, Sogawa, Hyvarinen, Kawahara,
  Washio, Hoyer, Bollen, and Hoyer]{shimizu2011directlingam}
Shohei Shimizu, Takanori Inazumi, Yasuhiro Sogawa, Aapo Hyvarinen, Yoshinobu
  Kawahara, Takashi Washio, Patrik~O Hoyer, Kenneth Bollen, and Patrik Hoyer.
\newblock Directlingam: A direct method for learning a linear non-gaussian
  structural equation model.
\newblock \emph{Journal of Machine Learning Research-JMLR}, 12\penalty0
  (Apr):\penalty0 1225--1248, 2011.

\bibitem[Spirtes et~al.(2000)Spirtes, Glymour, and
  Scheines]{spirtes2000causation}
Peter Spirtes, Clark~N Glymour, and Richard Scheines.
\newblock \emph{Causation, prediction, and search}.
\newblock MIT press, 2000.

\bibitem[Tikka et~al.(2023)Tikka, Helske, and Karvanen]{tikka2023clustering}
Santtu Tikka, Jouni Helske, and Juha Karvanen.
\newblock Clustering and structural robustness in causal diagrams.
\newblock \emph{Journal of Machine Learning Research}, 24, 2023.

\bibitem[Trefethen and Bau(2022)]{trefethen2022numerical}
Lloyd~N Trefethen and David Bau.
\newblock \emph{Numerical linear algebra}, volume 181.
\newblock Siam, 2022.

\bibitem[Wahl et~al.(2023)Wahl, Ninad, and Runge]{wahl2023foundations}
Jonas Wahl, Urmi Ninad, and Jakob Runge.
\newblock Foundations of causal discovery on groups of variables.
\newblock \emph{arXiv preprint arXiv:2306.07047}, 2023.

\bibitem[Wu et~al.(2024)Wu, Geiger, Icard, Potts, and
  Goodman]{wu2024interpretability}
Zhengxuan Wu, Atticus Geiger, Thomas Icard, Christopher Potts, and Noah
  Goodman.
\newblock Interpretability at scale: Identifying causal mechanisms in alpaca.
\newblock \emph{Advances in Neural Information Processing Systems}, 36, 2024.

\bibitem[Zennaro(2022)]{zennaro2022abstraction}
Fabio~Massimo Zennaro.
\newblock Abstraction between structural causal models: A review of definitions
  and properties.
\newblock In \emph{UAI 2022 Workshop on Causal Representation Learning}, 2022.

\bibitem[Zennaro et~al.(2023)Zennaro, Dr{\'a}vucz, Apachitei, Widanage, and
  Damoulas]{zennaro2023jointly}
Fabio~Massimo Zennaro, M{\'a}t{\'e} Dr{\'a}vucz, Geanina Apachitei, W~Dhammika
  Widanage, and Theodoros Damoulas.
\newblock Jointly learning consistent causal abstractions over multiple
  interventional distributions.
\newblock In \emph{2nd Conference on Causal Learning and Reasoning}, 2023.

\end{thebibliography}
\end{document}